\documentclass[11pt]{article}
\usepackage[table]{xcolor}
\usepackage[preprint]{acl}

\usepackage[utf8]{inputenc}
\usepackage{newunicodechar}

\usepackage{kotex}

\usepackage{times}
\usepackage{siunitx}
\usepackage{latexsym}
\usepackage{tablefootnote}
\usepackage{threeparttable}
\usepackage{pifont}
\usepackage{verbatim}

\usepackage{tcolorbox}
\usepackage{array}
\usepackage{ragged2e}
\usepackage{caption}
\usepackage[T1]{fontenc}
\usepackage[utf8]{inputenc}
\usepackage[verbose]{microtype}
\usepackage{inconsolata}
\usepackage{amsmath}
\usepackage{array}
\usepackage{multirow}
\usepackage{adjustbox}
\usepackage{graphicx}

\usepackage{booktabs}
\usepackage{url}
\usepackage{multicol}
\usepackage{ulem} 
\usepackage{makecell}
\usepackage{tabularx}
\usepackage{color}
\usepackage{soul}
\usepackage{lscape}
\usepackage{enumitem}
\usepackage{bbm}

\title{EXIT: Context-Aware Extractive Compression for Enhancing Retrieval-Augmented Generation}

\author{
  \textbf{Taeho Hwang}$^{1}$ \quad 
  \textbf{Sukmin Cho}$^{1}$ \quad 
  \textbf{Soyeong Jeong}$^{2}$ \quad 
  \textbf{Hoyun Song}$^{1}$ \\
  \quad \textbf{SeungYoon Han}$^{1}$ \quad 
  \textbf{Jong C. Park}$^{1}$\thanks{\;Corresponding author} \\
  $^{1}$School of Computing, $^{2}$Graduate School of AI \\
  Korea Advanced Institute of Science and Technology \\
  \texttt{\{doubleyyh,nelllpic,starsuzi,hysong,seungyoonee,jongpark\}@kaist.ac.kr}
}

\begin{document}
\maketitle
\begin{abstract}
We introduce \textbf{EXIT}, an extractive context compression framework that enhances both the effectiveness and efficiency of retrieval-augmented generation (RAG) in question answering (QA). Current RAG systems often struggle when retrieval models fail to rank the most relevant documents, leading to the inclusion of more context at the expense of latency and accuracy. While abstractive compression methods can drastically reduce token counts, their token-by-token generation process significantly increases end-to-end latency. Conversely, existing extractive methods reduce the latency but rely on independent, non-adaptive sentence selection, failing to fully utilize contextual information. EXIT addresses these limitations by classifying sentences from retrieved documents—while preserving their contextual dependencies—enabling parallelizable, context-aware extraction that adapts to query complexity and retrieval quality. Our evaluations on both single-hop and multi-hop QA tasks show that EXIT consistently surpasses existing compression methods and even uncompressed baselines in QA accuracy, while also delivering substantial reductions in inference time and token count. By improving both effectiveness and efficiency, EXIT provides a promising direction for developing scalable, high-quality QA solutions in RAG pipelines. Our code is available at \url{https://github.com/ThisIsHwang/EXIT}.
 
\end{abstract}
\section{Introduction}


Retrieval-Augmented Generation (RAG) \cite{RAG, DBLP:conf/iclr/KhandelwalLJZL20} is the task of enhancing the responses of Large Language Models (LLMs) with relevant external contexts or documents. By grounding answers in evidence, RAG systems have gained much attention for mitigating hallucination issues \cite{ralm, halueval} and improving factual reliability \cite{adaptive-rag, DBLP:conf/emnlp/XiaZLZLLZY24}.

However, they face significant challenges in practical deployment, as retrieval models sometimes fail to rank the most relevant documents at the top \cite{bm25, contriever}. One potential solution is to retrieve a larger set of documents to ensure the coverage of the necessary information, but this approach compromises both effectiveness and efficiency.
Specifically, for effectiveness, LLMs often struggle with processing long contexts, overlooking critical information located in the middle of contexts \cite{lostinthemiddle}. Additionally, irrelevant information in retrieved documents can act as distractors, significantly degrading the overall QA performance \cite{distractor, distractor2, distractor3}.
From an efficiency perspective, increasing the context size raises inference latency—due to quadratic complexity in attention computation \cite{inferencetime}—and API costs tied to input length\footnote{\url{https://openai.com/api/pricing/}}. Moreover, the context window limitations inherent in LLM architectures set strict upper bounds on the maximum input size~\cite{YaRN}.


\begin{figure}
    \centering
    \includegraphics[width=\columnwidth]{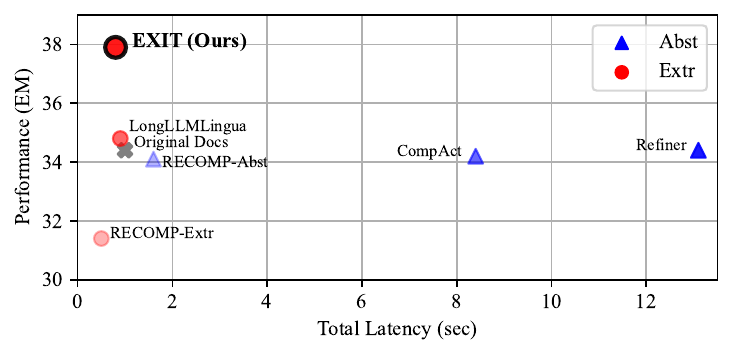}
\caption{Average QA accuracy (EM) and efficiency (Total Latency) across various compression methods using Contriever-MSMARCO as the retriever and Llama-3.1-8b-Instruct as the reader. Experiments were conducted on a single A100-80GB GPU, and latency (in seconds) was averaged over all samples from these datasets.} \label{fig:fig1}
    \vspace{-0.2in}
\end{figure}

To address these challenges, context compression has emerged as a promising solution, condensing essential information from multiple retrieved contexts through either abstractive or extractive approaches. While they can reduce inference time and filter out irrelevant information, both approaches still have notable drawbacks. Specifically, abstractive compression methods—often implemented via autoregressive generation—summarize or rewrite documents into a single condensed passage \cite{selective, recomp, compressor, Refiner, filco}, significantly increasing end-to-end latency due to their token-by-token generation process. For instance, as illustrated in Figure \ref{fig:fig1}, CompAct \cite{compressor}, a representative abstractive approach, takes over 8 seconds to process just five documents for a single query, whereas using the original document without compression takes only 1 second.

On the other hand, extractive compression approaches can offer a more efficient alternative \cite{recomp, llmlingua, longllmlingua}, by selecting relevant textual segments (e.g., sentences, or even token-level excerpts) directly from retrieved documents. This strategy reduces both compression time and overall latency. 
However, current extractive methods have yet to reach their full potential in terms of effectiveness \cite{decontextualization, llmlingua2}. They often rely on rigid selection criteria that do not adapt well to variations in query complexity or the quality of retrieved documents, and they frequently neglect to fully leverage the broader context when choosing which tokens or sentences to retain. Specifically, as illustrated in Figure \ref{fig:fig1}, while extractive approaches such as RECOMP-Extr \cite{recomp} achieve minimal compression time, their inability to dynamically adjust selection processes results in suboptimal QA performance.

\begin{figure*}[]
\centerline{\includegraphics[width=\linewidth]{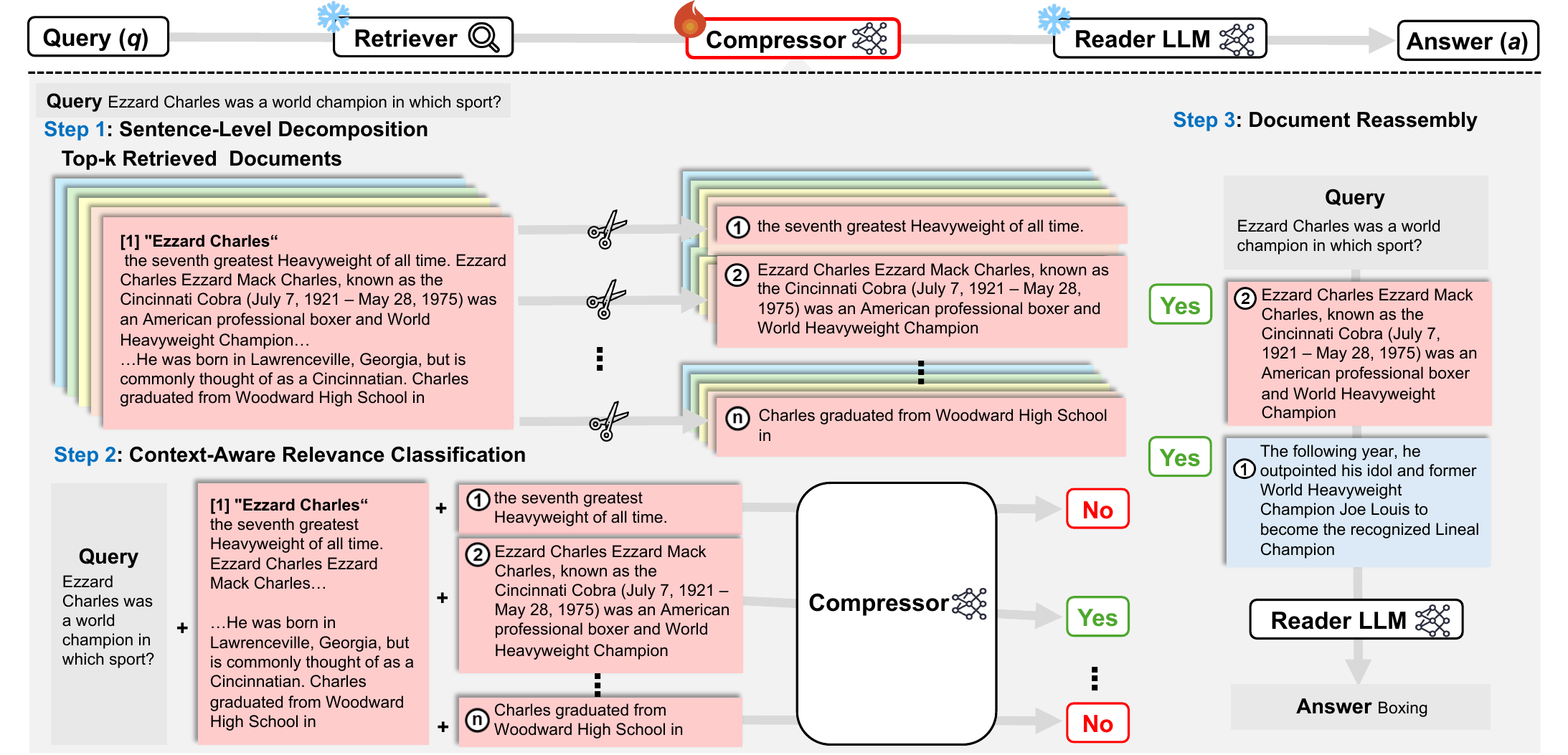}}
    \caption{Overview of our framework. First, the retrieved document is split into sentences. Next, each sentence is classified as either ``Yes" or ``No" using the Compressor. Finally, sentences with scores above the threshold are recombined in their original order to complete the compression.} \label{fig:fig2}
    \vspace{-0.2in}
\end{figure*}

Therefore, in this work, we propose a novel compression framework for RAG, \textbf{EX}tract\textbf{I}ve Contex\textbf{T} compression (EXIT), designed to enhance both effectiveness and efficiency by improving efficiency through an extractive compression and by enhancing effectiveness through dynamic, context-aware sentence selection. Specifically, as shown in Figure~\ref{fig:fig2}, EXIT operates in three stages: (1) splitting retrieved documents into sentences, (2) performing parallelizable binary classification (``Yes'' or ``No'') on each sentence to assess its relevance while considering its full document context, rather than evaluating them independently, and (3) recombining selected sentences while preserving their original order. Therefore, as shown in Figure~\ref{fig:fig1}, EXIT frames context compression as a sentence classification problem, enabling it to outperform both compression methods and the uncompressed baseline in terms of speed. Specifically, it reduces processing time from several seconds to about 1 second. Moreover, by leveraging context-aware and adaptive sentence selection, EXIT also surpasses other extractive methods in accuracy.
Also, we note that EXIT operates as a plug-and-play module that can be seamlessly integrated into any existing RAG pipeline without architectural modifications.

We evaluate EXIT on both single-hop QA tasks (NQ \cite{NQ}, TQA \cite{TQA}) and multi-hop QA tasks (HQA \cite{HQA}, 2WikiMultiHopQA \cite{2WIKI}). Experimental results demonstrate that EXIT not only improves effectiveness over both abstractive and extractive compression baselines but also significantly reduces latency compared to abstractive methods and the uncompressed baseline.


Our contributions are as follows:
\begin{itemize}

\item We identify and address the key weaknesses of existing context compression methods: abstractive methods incur prohibitive latency, while traditional extractive methods rely on rigid, non-adaptive content selection.


\item We propose \textbf{EXIT} (\textbf{EX}tract\textbf{I}ve Contex\textbf{T} Compression), an extractive compression framework that dynamically adjusts to query complexity and retrieval quality. 


\item We demonstrate, through extensive experiments, that EXIT surpasses previous compression methods and uncompressed retrievals, improving QA performance while significantly reducing both token counts and end-to-end latency.
\end{itemize}

\section{Related Work}

\noindent\textbf{Retrieval-Augmented Generation.~} RAG enhances LLMs by grounding responses in retrieved external documents~\cite{RAG, replug, ralm}.  However, increasing retrieved content often leads to longer contexts, higher inference costs~\cite{inferencetime},  and retrieval noise~\cite{distractor, lostinthemiddle}, impacting efficiency and accuracy.

To mitigate these challenges, ad-hoc methods have been proposed at different stages of the retrieval pipeline, including pre-retrieval and post-retrieval methods. Pre-retrieval methods, such as Landmark Embedding~\cite{landmark_embedding} and Late Chunking~\cite{late_chunking}, improve document representations to enhance retrieval effectiveness but lack adaptive filtering mechanisms after retrieval and may be difficult to integrate orthogonally with existing retrieval methods. Post-retrieval methods, including reranking~\cite{bert-rerank, llm-rerank, llara} and context compression~\cite{recomp, compressor, Refiner, longllmlingua}, refine retrieved documents through reordering or pruning. However, most post-retrieval methods overlook query complexity and operate on a fixed number of retrieved items, limiting their adaptability in balancing effectiveness and efficiency.

We introduce EXIT, a novel post-retrieval compression framework that adaptively filters query-relevant sentences while discarding redundancy. Unlike existing methods, EXIT dynamically adjusts to query complexity and retrieval quality, balancing efficiency and effectiveness. Its lightweight, adaptive design seamlessly integrates into RAG pipelines while remaining orthogonal to pre-retrieval methods.
\noindent\textbf{Context Compression.~} 
Context compression is a practical remedy for handling increasingly extensive prompt lengths in RAG pipelines. Existing methods commonly fall into soft or hard compression. Soft compression focuses on embedding prompts into compact continuous representations \cite{cc, gist, icae, autocompressor, xRAG}, but requires extensive training and architectural changes, making it unsuitable for black-box LLMs.
Hard compression, by contrast, removes non-essential textual content directly \cite{selective-context, llmlingua}, offering a plug-and-play solution even compatible with API-based models such as ChatGPT \cite{GPT-4_technical_report}. 

Hard compression are further divided into abstractive and extractive methods. Abstractive methods employ autoregressive models to generate query-focused summaries, thus reducing token counts at the cost of additional latency and potential hallucinations \cite{summary_halluciation}. For example, RECOMP-Abst \cite{recomp} uses a T5-based summarizer for token reduction but requires dataset-specific training and slows inference. CompAct \cite{compressor} and Refiner \cite{Refiner} take this method further by leveraging even larger LLMs with 7B parameters, compounding latency issues and increasing resource demands.
Extractive methods, such as RECOMP-Extr \cite{recomp}, directly select salient segments (e.g., sentences or tokens) from retrieved documents. This avoids the autoregressive bottleneck and mitigates hallucinations. However, their static and context-agnostic selection of only a few sentences per document limits its performance. Similarly, token-level methods such as LLMLingua family \cite{selective-context, llmlingua, llmlingua2, longllmlingua} can distort semantic coherence by removing key entities or splitting essential facts. 

In short, abstractive methods offer strong compression but suffer from latency and potential hallucinations, while extractive methods are often rigid and lack context awareness. Our work addresses these limitations by proposing a parallelizable, context-aware extractive compression framework. It adaptively selects sentences at scale, optimizing accuracy and speed in complex retrieval scenarios.

\section{Method}

In this section, we first present our problem formulation, the RAG pipeline with a compression stage, and our novel compression framework, \textbf{EXIT}, which is designed to extract key evidence for answering in a parallel manner.



\subsection{Problem Formulation}

\noindent\textbf{RAG Pipeline with Compression.} Given a query $q$ and a document corpus $\mathcal{C}$, a RAG pipeline first retrieves Top-$k$ relevant document set $D$:
\begin{equation}
\small
D = \{d_1, \dots, d_k\} = \texttt{Retriever}(q, \mathcal{C}),
\label{eq:retrieval}
\end{equation}
The retrieved documents within the document set $D$ are then processed by a compression module that preserves query-relevant information while significantly reducing input length:
\begin{equation}
\small
D' = \texttt{Compressor}(q, D) \; \text{s.t.}\; l(D') \ll l(D),
\end{equation}

where $l(\cdot)$ represents the function calculating the number of tokens in the document set. After compression, the number of tokens included in \(D'\) is substantially decreased compared to \(D\). Finally, an LLM generates the answer $a$ using the compressed set $D'$ and the given query $q$:
\begin{equation}
\small
a = \texttt{LLM}(q, D').
\label{eq:generation}
\end{equation}


\noindent\textbf{Objectives of Compression.}
Effective compression in the RAG pipeline must satisfy three key criteria: (1) \textbf{Token Reduction}–$D'$ should have fewer tokens to speed up answer generation; (2) \textbf{Retention of Key Evidence}–essential information must be preserved to maintain answer accuracy; and (3) \textbf{Efficient Processing}–compression should be fast enough to avoid introducing significant latency.

While token count influences the reader’s reading time, it alone does not capture the full latency of the pipeline. In practice, end-to-end latency also includes the time spent on the compression step itself. Therefore, an efficient RAG system must minimize both the size of the input and the total time taken from retrieval to answer generation.


\subsection{Extractive Context Compression (EXIT)}
To achieve three objectives of the compression step, EXIT consists of three main components: sentence-level decomposition, context-aware relevance classification, and document reassembly. Importantly, EXIT functions as a \textbf{plug-and-play module} that integrates seamlessly into existing RAG pipelines, independent of the specific retriever or reader model used.

\noindent\textbf{Sentence-Level Decomposition.}
In Step 1 of Figure~\ref{fig:fig2}, EXIT divides each retrieved document into individual sentences using a rule-based sentence tokenizer. For each document \(d_i\in D\), it produces a sentence set \(S_i = \{s_{i1}, s_{i2}, \dots, s_{in}\}\), where \(s_{ij}\) is the \(j\)-th sentence in document \(i\). Operating at the sentence level avoids the fragmentation of key phrases and preserves entity relationships that token-level compression techniques \cite{llmlingua} often disrupt. As a result, the compressed context preserves both syntactic coherence and semantic integrity, ensuring that key information is retained.

\noindent\textbf{Context-Aware Relevance Classification.}
To efficiently filter sentences in $D$ that contain key evidence for answering a query, we introduce a relevance evaluation method based on \textbf{context awareness} and \textbf{single-token prediction}. First, incorporating the entire document $d_i$ as context is essential, as understanding a sentence often requires the broader document context rather than an isolated sentence. This ensures that no relevant information is overlooked and enables more effective compression. Second, to maintain efficiency, we adopt lightweight single-token prediction instead of multi-token generation \cite{compressor, Refiner}, which can introduce computational overhead. Inspired by~\citet{unieval}, we leverage probabilistic classification for sentence relevance assessment. Given  query $q$, document $d_i$, and sentence $s_{ij}$, the evaluation model predicts relevance using a binary classification with ``Yes" and ``No" labels:
\begin{equation}
\small
    r_{ij} = \frac{P(\text{``Yes''}| q,d_i,s_{ij})}{P(\text{``Yes''}| q,d_i,s_{ij}) + P(\text{``No''}| q,d_i,s_{ij})}
\end{equation}
where $P(\cdot|\cdot)$ represents the likelihood from the evaluation model. This computation is parallelized across sentences for efficiency.  

EXIT then selects sentences with a relevance score above a predefined threshold $\tau$, ensuring that only key information is retained. Unlike fixed-size selection, our framework produces an \textbf{adaptive number of sentences} in the compressed set $D'$, aligning with prior work \cite{adaptive-rag} that acknowledges query-dependent information needs. This approach optimizes compression while preserving essential evidence.

\noindent\textbf{Document Reassembly.} As shown in Step 3 of Figure~\ref{fig:fig2}, EXIT reconstructs the compressed document $D'$ by concatenating selected sentences in their original order, preserving coherence and logical flow~\cite{dslr} for accurate downstream reasoning.

\subsection{Classifier Model Training}
\label{sec:training}
\noindent\textbf{Training Strategy.}
Our goal is to train a relevance classifier capable of accurately identifying which sentences provide the evidence required to answer a query. To approximate real-world complexity, we utilize a question-answering dataset that requires multi-sentence reasoning and offers explicit sentence-level annotations of essential information. Leveraging these annotations, we model three typical retrieval outcomes: (1) sentences containing necessary evidence, (2) seemingly relevant sentences missing key details, and (3) entirely irrelevant sentences.

\noindent\textbf{Data Sampling.}
To train a robust relevance classifier, we construct a diverse dataset reflecting various post-retrieval conditions. Positive samples include sentences essential for correct answers, while hard negatives consist of remaining sentences from the same documents. Additionally, random negatives pair queries with unrelated sentences, helping the model distinguish relevance from noise. This balanced sampling enables effective filtering of non-essential information without relying on explicit supervision.

\noindent\textbf{Training Procedure.}  
Each training instance is represented as \((q, s, d, l)\), where \(q\) is the query, \(s\) is a candidate sentence, \(d\) is the document containing \(s\), and \(l \in \{\text{“Yes”}, \text{“No”}\}\) indicates whether \(s\) provides the required evidence. We employ a binary cross-entropy loss function to train the classifier:
\begin{equation}
\small
    \mathcal{L} = -\mathbbm{1}_{l=\text{“Yes”}}\log P(\text{“Yes”}) - (1-\mathbbm{1}_{l=\text{“Yes”}})\log P(\text{“No”}),
\end{equation}

By exposing the classifier to a balanced and diverse set of retrieval scenarios, we improve its ability to generalize and reliably identify sentences that contain the critical evidence for answering queries.

\section{Experiment Setups}
We conduct comprehensive experiments to evaluate EXIT's effectiveness and efficiency in context compression for RAG systems. More implementation details of our experiments are in Appendix~\ref{app:implementation}.

\noindent\textbf{Datasets.~}
We evaluate on both single-hop and multi-hop question answering datasets: \textbf{NaturalQuestions (NQ)}~\cite{NQ} and \textbf{TriviaQA (TQA)}~\cite{TQA} for single-hop QA; \textbf{HotpotQA (HQA)}~\cite{HQA} and \textbf{2WikiMultihopQA (2WIKI)}~\cite{2WIKI} for multi-hop QA. We use the test set for TQA evaluations and development sets for all other datasets. For the train dataset for the classifier, we exploit the train split of HQA, which has relevant annotations to each sentence in the multiple documents for a query.

\noindent\textbf{Model Configuration.~}
Our system consists of three primary components. The retriever employs \textbf{Contriever-MSMARCO}~\cite{contriever}, a dense retriever fine-tuned on MSMARCO~\cite{msmarco}. The EXIT compressor utilizes \textbf{Gemma-2B-it}~\cite{gemma2}, optimized for efficient parallel processing. For the reader model, we exploit two scales of instruction-tuned models: \textbf{Llama3.1-\{8, 70\}B-Instruct}~\cite{llama3}.


\begin{table*}[]
    \centering
    \caption{Performance across models and datasets, measured by EM, F1, and inference latency (Lat.). 8B reader experiments were conducted on a single A100-80GB GPU, while 70B reader experiments utilized 4 A100-80GB GPUs in parallel. Best results for each dataset are highlighted in \textbf{bold}, and second best results are \underline{underlined}. The ``Type'' column denotes whether a given compressor is abstractive (Abs.) or extractive (Ext.).}
    \small
    \label{tab:performance_comparison}
    \setlength{\tabcolsep}{5pt}
    \begin{adjustbox}{width=\textwidth}
    \begin{tabular}{l|c|ccc|ccc|ccc|ccc|ccc}
        \toprule[1.5pt]
        \multirow{2}{*}{\textbf{Compressor}} & \multirow{2}{*}{\textbf{Type}} 
        & \multicolumn{3}{c|}{\textbf{NQ}} 
        & \multicolumn{3}{c|}{\textbf{TQA}} 
        & \multicolumn{3}{c|}{\textbf{HQA}} 
        & \multicolumn{3}{c|}{\textbf{2WIKI}}
        & \multicolumn{3}{c}{\textbf{AVG.}} \\
        \cmidrule[0.8pt]{3-17}
        &  & \scriptsize \textbf{EM} $\uparrow$ & \scriptsize \textbf{F1} $\uparrow$ & \scriptsize \textbf{Lat.} $\downarrow$ 
        & \scriptsize \textbf{EM} $\uparrow$ & \scriptsize \textbf{F1} $\uparrow$ & \scriptsize \textbf{Lat.} $\downarrow$ 
        & \scriptsize \textbf{EM} $\uparrow$ & \scriptsize \textbf{F1} $\uparrow$ & \scriptsize \textbf{Lat.} $\downarrow$ 
        & \scriptsize \textbf{EM} $\uparrow$ & \scriptsize \textbf{F1} $\uparrow$ & \scriptsize \textbf{Lat.} $\downarrow$
        & \scriptsize \textbf{EM} $\uparrow$ & \scriptsize \textbf{F1} $\uparrow$ & \scriptsize \textbf{Lat.} $\downarrow$ \\
        \midrule[1pt]
        \multicolumn{17}{c}{\cellcolor{gray!10}\textbf{Llama3.1-8B-Instruct}} \\
        \midrule[0.8pt]
        Original Docs & - & \underline{34.6} & \underline{47.1} & 1.0 & 58.8 & 68.6 & 0.9 & 28.1 & 38.6 & 1.0 & 16.1 & 24.9 & 1.1 & 34.4 & \underline{44.8} & 1.0 \\
        RECOMP-Abst & Abs. & 31.3 & 43.2 & 1.6 & 55.9 & 65.7 & 1.4 & 26.5 & 37.0 & 2.2 & \underline{22.7} & \underline{29.1} & 2.1 & 34.1 & 43.7 & 1.8 \\
        CompAct & Abs. & 32.9 & 44.6 & 8.5 & 58.1 & 67.7 & 8.8 & \underline{28.8} & 39.8 & 8.3 & 16.8 & 26.0 & 8.1 & 34.2 & 44.5 & 8.4 \\
        Refiner & Abs. & 32.9 & 45.0 & 28.1 & 59.2 & \underline{68.9} & 10.9 & \underline{28.8} & \underline{40.0} & 6.9 & 16.8 & 25.4 & 6.4 & 34.4 & \underline{44.8} & 13.1 \\
        RECOMP-Extr & Ext. & 34.6 & 44.6 & \textbf{0.5} & 56.5 & 65.1 & \textbf{0.4} & 23.4 & 32.8 & \textbf{0.4} & 11.2 & 19.6 & \textbf{0.6} & 31.4 & 40.5 & \textbf{0.5} \\
        LongLLMLingua & Ext. & 30.2 & 41.5 & 0.9 & \underline{59.4} & 68.0 & 0.8 & 28.0 & 38.0 & \underline{0.8} & 21.5 & 27.4 & \underline{0.9} & \underline{34.8} & 43.7 & 0.9 \\
        \rowcolor{blue!5} EXIT (Ours) & Ext. & \textbf{35.9} & \textbf{47.8} & \underline{0.8} & \textbf{60.8} & \textbf{69.9} & \underline{0.7} & \textbf{30.6} & \textbf{41.5} & \underline{0.8} & \textbf{24.2} & \textbf{30.8} & \underline{0.9} & \textbf{37.9} & \textbf{47.5} & \underline{0.8} \\
        \midrule[1pt]
        \multicolumn{17}{c}{\cellcolor{gray!10}\textbf{Llama-3.1-70B-Instruct}} \\
        \midrule[0.8pt]
        Original Docs & - & 35.6 & 48.0 & 8.6 & 65.1 & 73.9 & 7.7 & 33.7 & 44.5 & 8.3 & 20.8 & 28.3 & 9.1 & 38.8 & 48.7 & 8.4 \\
        RECOMP-Abst & Abs. & 34.1 & 47.0 & 4.5 & 61.3 & 70.6 & 3.3 & 30.3 & 40.8 & 4.4 & 24.2 & 30.3 & 4.2 & 37.5 & 47.2 & 4.1 \\
        CompAct & Abs. & 34.1 & 45.4 & 11.9 & 62.6 & 71.1 & 11.7 & 33.8 & 44.1 & 11.0 & 20.5 & 27.4 & 11.6 & 37.8 & 47.0 & 11.5 \\
        Refiner & Abs. & 35.3 & \underline{47.1} & 42.5 & 64.3 & 73.0 & 18.3 & 33.8 & 44.7 & 14.6 & 21.2 & 28.0 & 11.2 & 38.7 & 48.2 & 21.6 \\
        RECOMP-Extr & Ext. & \underline{35.8} & 45.3 & \textbf{2.5} & 63.5 & 71.0 & \textbf{2.2} & 27.6 & 36.7 & \textbf{2.9} & 13.8 & 19.3 & \textbf{3.3} & 35.2 & 43.1 & \textbf{2.7} \\
        LongLLMLingua & Ext. & 32.2 & 44.0 & 4.4 & \underline{66.7} & \underline{75.2} & 3.9 & \underline{34.1} & \underline{45.3} & 4.0 & \underline{28.3} & \textbf{34.8} & 4.3 & \underline{40.3} & \underline{49.8} & 4.1 \\
        \rowcolor{blue!5} EXIT (Ours) & Ext. & \textbf{36.9} & \textbf{49.4} & \underline{3.9} & \textbf{67.3} & \textbf{75.9} & \underline{3.1} & \textbf{37.0} & \textbf{48.3} & \underline{3.3} & \textbf{28.6} & \underline{34.5} & \underline{3.5} & \textbf{42.5} & \textbf{52.0} & \underline{3.5} \\
        \bottomrule[1.5pt]
    \end{tabular}
    \end{adjustbox}
\end{table*}

\noindent\textbf{Baselines.~}
We compare EXIT against the following context compression methods: \noindent\textbf{1) Original Documents} serves as uncompressed baseline. For abstractive methods, \noindent\textbf{2) RECOMP-Abs} \cite{recomp} uses T5-based (775M) summarization tuned for NQ, TQA, HQA (HQA model for 2WIKI), \noindent\textbf{3) CompAct} \cite{compressor} implements Mistral-7B-based iterative compression with 5-segment blocks, and \noindent\textbf{4) Refiner}~\cite{Refiner} uses Llama2-7B-based compression. For extractive methods, \noindent\textbf{5) RECOMP-Extr} \cite{recomp} employs Contriever-based (110M) sentence-level extraction tuned for NQ, TQA, HQA (HQA model for 2WIKI), and \noindent\textbf{6) LongLLMLingua} \cite{longllmlingua} uses Llama2-7B-chat for token-level extraction with 0.4 dynamic compression rate.

\noindent\textbf{Evaluation Metrics.~}
We use three metrics: \textbf{Exact Match (EM)} and \textbf{F1} score to measure effectiveness in question answering, and \textbf{end-to-end inference latency (Lat.)} in seconds to assess efficiency. End-to-end latency captures the total wall-clock time required to generate an answer for a single query, including both compression time and LLM inference time. We report the average latency per sample, computed over the entire evaluation set. This provides a more holistic efficiency metric than token count alone, as it directly reflects the practical runtime performance of the full RAG pipeline.

\noindent\textbf{Implementation Details.~}
We perform document retrieval using the December 2018 Wikipedia dump and apply SpaCy for sentence segmentation. The LLM outputs are generated with a temperature of 0.0 and a top-$P$ of 1.0. We set the relevance threshold $\tau$ to 0.5 based on empirical tuning. All experiments are conducted on A100-SXM4-80GB GPUs.



\begin{figure*}[ht]
\centerline{\includegraphics[width=\linewidth]{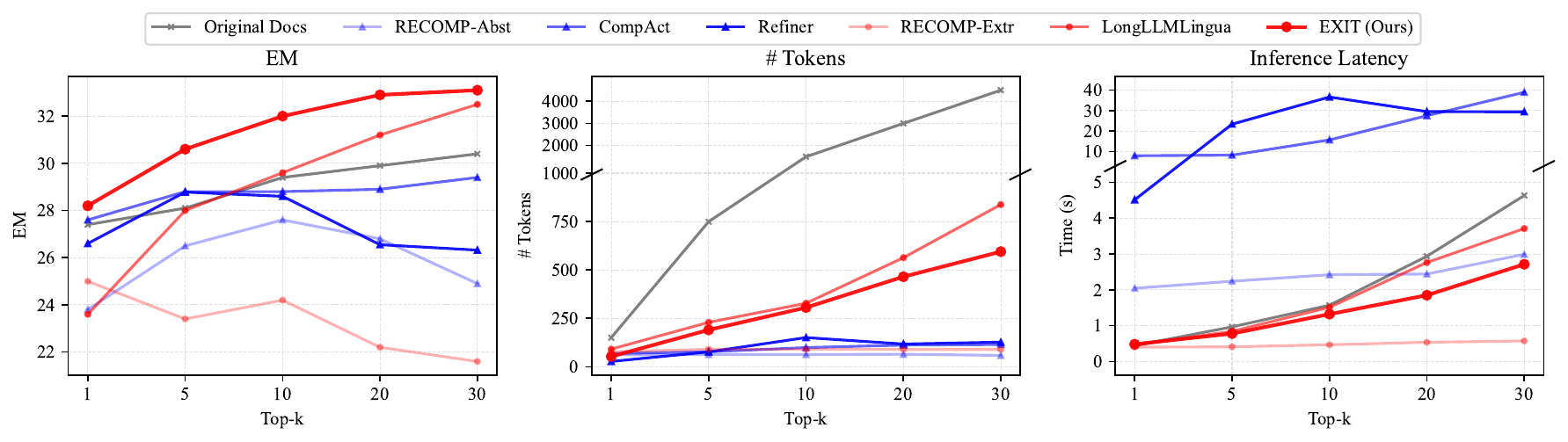}}
    \vspace{-0.2in}
    \caption{Performance analysis on HQA across different Top-\(k\) values (1, 5, 10, 20, 30), comparing accuracy, token retention, and inference latency between baselines and our method. All experiments were conducted on a single A100-80GB GPU.} \label{fig:topk_analysis}
    \vspace{-0.2in}
\end{figure*}
\section{Main Results}

Table~\ref{tab:performance_comparison} summarizes our evaluation results across multiple datasets and compression methods. With the 8B reader, EXIT demonstrates strong generalization: although trained solely on HQA, it effectively addresses both single-hop (NQ, TQA) and multi-hop (2WIKI) queries under out-of-domain conditions. Compared to all baseline methods, EXIT consistently improves EM scores—for instance, by 1.3 and 2.0 points on NQ and TQA, and by even larger margins of 2.5 and 8.1 points on HQA and 2WIKI, respectively. Notably, these accuracy gains come with an average latency of just 0.8s, substantially faster than abstractive compression methods.

The benefits of EXIT become more pronounced at larger scales. Using the 70B reader, EXIT surpasses the accuracy of all competing methods, averaging a 3.7-point improvement in EM and a 3.3-point improvement in F1 over the uncompressed baseline. On HQA, it achieves a 3.3-point EM gain while maintaining an efficient 3.5s latency—faster than using uncompressed documents and still competitive with the previously fastest method, RECOMP-Extr, but with significantly higher accuracy. EXIT's effectiveness and efficiency, especially with larger models, makes it a practical solution for large-scale QA applications.

\section{Analyses}
We conduct a series of analyses examining EXIT's robustness, classification performance, latency factors, and design choices under various configurations. Additional experimental results and analyses are provided in Appendix \ref{app:Experimental}.

\subsection{Robustness Analysis}
To examine EXIT’s robustness as the retrieval set size grows, we gradually increased the number of retrieved documents (\(k \in \{1, 5, 10, 20, 30\}\)) with an 8B reader, as shown in Figure~\ref{fig:topk_analysis}. We found that EXIT steadily improves EM scores—from 28.2 points at \(k=1\) to 33.1 points at \(k=30\)—while avoiding the performance degradation seen in RECOMP variants and Refiner at high \(k\) values. 
Also, we measured the impact on the efficiency of the RAG pipeline with token counts and end-to-end latency, confirming that EXIT significantly reduces context from 4,497.1 tokens to 594.4 tokens (86.8\% fewer) at \(k=30\), even improving the quality.
Also, EXIT's latency scales nearly linearly (0.48s to 2.71s) and is much faster than the abstraction methods and the uncompressed baseline. These results demonstrate that EXIT consistently delivers significant accuracy improvements with minimal inference costs, regardless of the number of documents, making it well-suited for tasks involving larger retrieval sets.

\subsection{Understanding End-to-End Latency Factors}
\begin{figure}[t]
\centerline{\includegraphics[width=\linewidth]{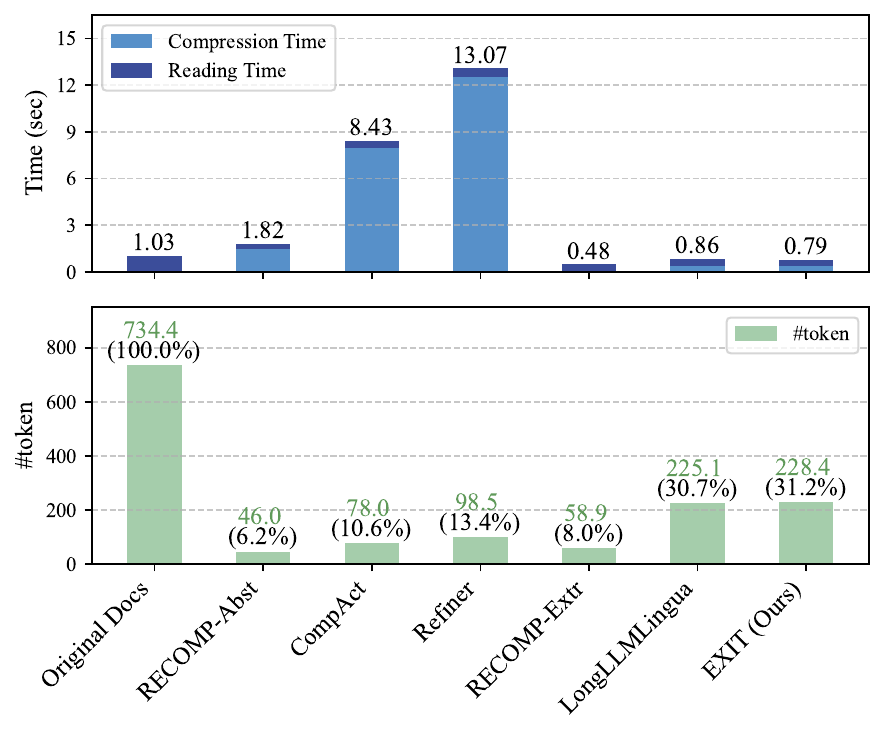}}
    \vspace{-0.2in}
    \caption{Comparison of compression and reading latency across baselines and our method in QA setting. Experiments were conducted on a single A100 GPU.} \label{fig:compression_analysis}
\end{figure}

While previous work has primarily focused on minimizing token counts to reduce reading time, we argue that \textbf{end-to-end latency}—including both compression and generation time—offers a more practical measure of efficiency for RAG pipelines.

Figure~\ref{fig:compression_analysis} presents a breakdown of total latency into compression time and reading time, along with the average token count of the compressed context. Methods such as RECOMP-Abst (6.2\% token ratio) and CompAct (10.6\%) achieve strong compression but suffer from long compression times (1.46s and 7.99s, respectively), resulting in overall latency that exceeds the uncompressed baseline. By contrast, EXIT achieves a balanced 31.2\% token ratio and fast compression (0.36s), reducing total latency to 0.79s—a 23.3\% improvement over the uncompressed baseline (1.03s).

We acknowledge that EXIT may retain more content when the question is simple or the retrieved documents are of high quality. However, our goal is not to maximize compression alone, but to \textbf{minimize total latency} while preserving answer accuracy. Aggressive compression in such cases could introduce unnecessary overhead or risk omitting useful context. EXIT's sentence-level classification balances this trade-off, allowing rapid, context-aware compression with consistent gains in both speed and accuracy, as shown in Table~\ref{tab:performance_comparison}.

This analysis highlights that \textbf{efficient compression is not solely about reducing tokens}—it is about achieving the best trade-off between compression time and model input size. EXIT is designed with this principle in mind, making it a practical solution for real-world RAG applications.

\subsection{Computational Complexity and Efficiency Trade-offs}
\begin{table}[t]
\centering
\caption{TFLOPs, latency, and relative speedup of compression methods. Latency is measured on a single A100-80GB GPU. Speedup is relative to the uncompressed baseline (Original Docs).}
\label{tab:complexity_latency}
\resizebox{\columnwidth}{!}{
\begin{tabular}{lcccc}
\toprule
\textbf{Method} & \textbf{Comp. TFLOPs} & \textbf{Read. TFLOPs} & \textbf{Latency (s)} & \textbf{Speedup} \\
\midrule
Original Docs     & --      & 13.04  & 0.95   & 1.00× \\
RECOMP-Abst       & 0.54    & 1.60   & 2.31   & 0.41× \\
CompAct           & 18.35   & 2.05   & 12.86  & 0.07× \\
Refiner           & 15.03   & 2.51   & 6.78   & 0.14× \\
RECOMP-Extr       & 0.27    & 1.84   & 0.39   & 2.43× \\
LongLLMLingua     & 46.39   & 5.06   & 0.82   & 1.16× \\
\textbf{EXIT (ours)} & \textbf{35.44} & \textbf{2.96} & \textbf{0.88} & \textbf{1.08×} \\
\bottomrule
\end{tabular}
}
\end{table}

To strengthen the justification of EXIT’s efficiency, we analyze not only latency but also computational complexity, measured in TFLOPs, using DeepSpeed’s FlopsProfiler~\cite{deepspeed}. Table~\ref{tab:complexity_latency} reports the total FLOPs required for compression and reading (inference), as well as the resulting latency, all measured on a single A100-80GB GPU.

Although EXIT incurs higher total TFLOPs (38.40) than some competing methods, its design enables parallel sentence-level classification, which significantly reduces end-to-end latency. By contrast, abstractive methods like CompAct and Refiner involve autoregressive decoding, which must be executed sequentially and is subject to memory I/O bottlenecks, leading to much longer runtimes (up to 12.86 seconds).

This reveals a key trade-off: EXIT utilizes more GPU computation, but this computation is efficiently executed in parallel. Since many LLM inference pipelines are memory-bound rather than compute-bound, EXIT takes advantage of underutilized compute capacity to reduce latency. For example, compared to the uncompressed baseline (0.95s), EXIT achieves a lower latency of 0.88s, despite higher FLOPs.

These results demonstrate that EXIT provides practical efficiency by converting increased computation into latency reduction through parallel execution, offering a better balance between accuracy and response time than autoregressive alternatives.

\subsection{Ablation Studies}

\begin{table}[]
   \centering
   \caption{Ablation studies on HQA examining (1) training data composition (Pos, H-Neg, Neg), (2) adaptive vs. fixed-length sentence selection, and (3) the impact of incorporating passage context during classification. 
   }
   \vspace{-0.1in}
   \label{tab:ablation}
   \setlength{\tabcolsep}{4.5pt}
   \renewcommand{\arraystretch}{1.1}
   \begin{adjustbox}{width=0.4\textwidth}
   \begin{tabular}{l|ccc}
       \toprule[1.5pt]
       \multirow{1}{*}{\textbf{Configuration}} 
       & \scriptsize \textbf{EM} $\uparrow$
       & \scriptsize \textbf{F1} $\uparrow$
       & \scriptsize \textbf{\# token} $\downarrow$ \\
       \midrule[1pt]
       \rowcolor{gray!10}
       Ours (Pos + H-Neg + Neg) & \textbf{31.6} & \textbf{42.6} & 195.1 \\
       \midrule[0.2pt]
       Pos + H-Neg & 30.0 & 41.3 & 286.8 \\
       Pos + Neg & 29.8 & 40.9 & 404.6 \\
       \midrule[0.2pt]
       w/o Adaptive Sentence Selection (2 sents) & 29.4 & 40.7 & \textbf{91.0} \\
       w/o Adaptive Sentence Selection (4 sents) & 30.2 & 41.4 & 166.5 \\
       \midrule[0.2pt]
       w/o Passage as context & 30.4 & 42.3 & 157.4 \\
       \bottomrule[1.5pt]
   \end{tabular}
   \end{adjustbox}
   \vspace{-0.1in}
\end{table}

To better understand how the design choices in EXIT affect its overall performance and efficiency, we conduct ablation studies focusing on three key components: data sampling strategy, adaptive sentence selection, and context-aware extraction. Table~\ref{tab:ablation} summarizes these results.

\noindent\textbf{Data Sampling Strategy.}
Our training data combines positive, hard negative, and random negative samples to mirror the diversity of real-world retrieval scenarios. Compared to using only subsets of these sample types, the comprehensive strategy improves EM by 1.6 points over using only hard negatives and by 1.8 points over only random negatives. Relying solely on positive and random negatives led to excessive token retention, while depending only on hard negatives diminished the model’s ability to filter out spurious retrieval noise.

\noindent\textbf{Adaptive Sentence Selection.}
We compare our adaptive selection mechanism to a fixed-length strategy, which limits selection to four sentences. While the fixed-length method reduces token count to 166.5, it lowers EM by 1.4 and F1 by 1.2 points. This highlights the importance of adapting sentence selection based on query complexity and document relevance rather than imposing a static constraint.

\noindent\textbf{Context-Aware Extraction.}
We evaluate the effect of considering full document context when determining sentence relevance. Removing surrounding context reduces token usage by 38 tokens but decreases EM by 1.2 points, highlighting the importance of contextual awareness in preserving answer accuracy. While context-aware extraction slightly increases token count, it ensures more precise and reliable responses.

These findings confirm that a balanced training data strategy enhances robustness, that adaptive sentence selection ensures efficiency, and that full document consideration preserves accuracy. 

\subsection{Compressor Model Size and Strategy Impact}  
\begin{figure}[]
\centerline{\includegraphics[width=\linewidth]{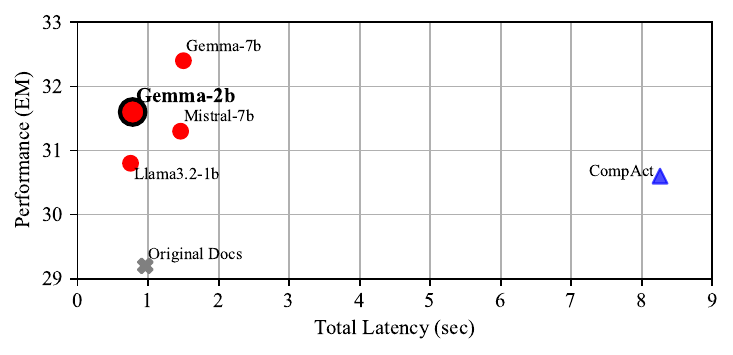}}
    \vspace{-0.2in}  
    \caption{Ablation on HQA comparing EM scores and latency for different model configurations within EXIT (red dot). CompAct and Original Docs are included as indicators.
    Experiments on a single A100-80GB GPU.} \label{fig:model_ablation}
\end{figure}

\noindent\textbf{Model Size Considerations.}  
Figure~\ref{fig:model_ablation} presents an ablation study examining how different base models influence EM scores and total latency. 
Note that all models trained within EXIT achieve superior accuracy compared to uncompressed baselines and fast compression under 2 seconds.
Specifically, Gemma-2B, our base classifier model, achieves a favorable balance between effectiveness and efficiency, delivering 31.6 EM points in 0.78s. 
When moving to the largest model, Gemma-7B, the highest accuracy (32.4 EM) is achieved, yet it also inflates latency to 1.50s, slightly exceeding the time of uncompressed documents. 
These results suggest that scaling up model parameters can improve performance but may also compromise latency benefits, emphasizing the flexibility of our proposed method by selecting an appropriate model depending on the user requirement.



\noindent\textbf{Abstractive vs. Extractive Compression.}  
Figure~\ref{fig:model_ablation} also compares our extractive approach, EXIT, against CompAct, a 7B-scale abstractive compressor. Using Mistral-7B as the base model for both methods, EXIT (31.3 EM, 1.46s) significantly outperforms CompAct (30.6 EM, 8.26s) in terms of latency and maintains competitive accuracy. This stark difference underscores that the compression strategy, not the just model size, heavily influences efficiency. By relying on extraction rather than iterative summarization, EXIT capitalizes on a large-scale model to preserve high accuracy without incurring prohibitively long inference time.

\begin{table}[t]
\centering
\small
\caption{Performance comparison across different retrieval and reranking methods on 2WIKI.}
\vspace{-0.1in}
\renewcommand{\arraystretch}{0.9} 
\begin{tabular}{lccc}
\toprule
\small
\textbf{Method} & \textbf{EM} & \textbf{F1} & \textbf{\#tok.} \\
\midrule
\multicolumn{4}{c}{\textbf{Passage-Level (Top-1)}} \\
\midrule
bge-m3 & 14.6 & 22.7 & 156.5 \\
bge-reranker-v2-gemma & 15.0 & 22.9 & 156.2 \\
RepLLaMA & 16.0 & 24.1 & 157.2 \\
UPR & 9.2 & 16.8 & 155.3 \\
Yes/No Ranking Prompt & 13.4 & 21.8 & 157.5 \\
Pairwise Ranking Prompt & 13.2 & 21.5 & 156.2 \\
\midrule
\multicolumn{4}{c}{\textbf{Sentence-Level (Top-5)}} \\
\midrule
bge-m3 & 16.6 & 24.2 & 216.9 \\
bge-reranker-v2-gemma & 18.4 & 26.2 & 203.2 \\
RepLLaMA & 14.8 & 23.2 & 239.0 \\
UPR & 7.4 & 14.6 & 148.9 \\
Yes/No ranking prompt & 14.6 & 21.8 & 135.6 \\
Pairwise ranking prompt & 11.6 & 19.4 & 209.9 \\
\midrule
Ours & \textbf{24.8} & \textbf{30.1} & \textbf{150.2} \\
\bottomrule
\end{tabular}
\label{tab:reranker}
\vspace{-0.1in}
\end{table}

\subsection{Comparison with Reranking Methods}

We compare EXIT with post-retrieval reranking methods in Table~\ref{tab:reranker}, including bge-m3~\cite{bge-m3}, bge-reranker-v2-gemma~\cite{llara}, RepLLaMA~\cite{repllama}, UPR~\cite{upr}, Yes/No ranking prompting~\cite{holistic}, and Pairwise ranking prompting~\cite{prp}. EXIT consistently achieves higher EM and F1 scores, demonstrating its ability to adaptively select query-relevant sentences while filtering out redundancy. By contrast, Top-$k$ re-ranking methods rely on static ranking, often retaining irrelevant content, increasing token count, and diminishing output quality. Moreover, they lack sentence-level context awareness, treating sentences independently without considering their broader document context, which can lead to incoherent or incomplete retrieval. These results underscore EXIT’s advantage in providing a more efficient and contextually relevant input for RAG than other post-retrieval methods.

\subsection{Leveraging LLMs for Pseudo-Annotation}
\begin{figure}[t]
    \centering
    \includegraphics[width=0.6\linewidth]{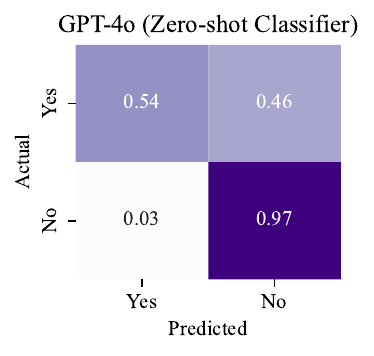}
    \caption{Confusion matrix of GPT-4o zero-shot classifier on HotpotQA. The model demonstrates high precision for identifying irrelevant sentences (``No'') but shows moderate recall for relevant ones (``Yes'').}
    \label{fig:gpt4o_confusion}
\end{figure}

To explore scalable alternatives to manual sentence-level annotations, we evaluated the use of a large language model (GPT-4o) as a zero-shot relevance classifier for generating pseudo-labels. Without any fine-tuning, GPT-4o achieved strong filtering precision, correctly identifying 97\% of irrelevant sentences, though with lower recall (54\%) for relevant ones (Figure~\ref{fig:gpt4o_confusion}). Despite this, EXIT trained with GPT-4o annotations still delivered competitive QA performance (Table~\ref{tab:exit_vs_baselines}), highlighting the feasibility of LLM-based pseudo-annotation as a scalable supervision strategy. These findings suggest a promising direction for improving generalization across domains by leveraging LLM-generated training data beyond curated datasets like HQA.

\section{Conclusion}
We present EXIT, an efficient context compression framework for RAG systems that leverages parallel processing and context-aware, adaptive sentence selection. Our experiments demonstrate that EXIT achieves superior performance across both single-hop and multi-hop QA tasks while maintaining practical inference speeds. Despite being trained only on HQA, EXIT shows a strong zero-shot generalization ability and proves effective across a wide range of open-source models of varying sizes. These results show that parallel extraction with smaller models outperforms larger abstractive methods, making it practical for real-world RAG.


\normalem

\section*{Limitation}
This work focuses on RAG pipelines, and its effectiveness in general long-context scenarios like LongBench~\cite{longbench} remains future work. Our approach relies on explicit sentence-level annotations for training, which were manually obtained in our experiments. While these could be automated using GPT-4 supervision or reader-derived signals (as discussed in Section~\ref{app:signal}), we have not yet explored fully automated annotation strategies—an important direction for future work.

EXIT performs sentence-level extraction, which may introduce limitations when sentences are overly long, noisy, or contain irrelevant details. While our context-aware classifier helps mitigate this, some ambiguity may persist, especially in cases involving complex or entity-heavy sentences. Additionally, although preserving the original sentence order empirically yields the best performance, we observed that removing intermediate sentences can occasionally lead to unnatural or incoherent reconstructions. Addressing this may require future improvements such as sentence decomposition, coherence-aware selection, or lightweight sentence refinement mechanisms.

We also focus on a single-step RAG setting, excluding iterative or recursive retrieval~\cite{iterativeretrieval, recursiveretrieval, dsp}. However, EXIT is orthogonal to these approaches and can be integrated by compressing retrieved content at each step.

Finally, while EXIT avoids the latency bottlenecks of autoregressive generation, further efficiency gains could be achieved through architectural optimizations such as prefix caching or RadixAttention~\cite{sglang}, which remain open areas for future exploration.

\section*{Ethics Statement}
This work enhances RAG-based QA without generating new content beyond what is retrieved. However, biases and inaccuracies in the source documents can still propagate through our compression process. Ensuring the reliability, fairness, and proper curation of underlying corpora is essential for ethical deployment. Future efforts should integrate bias detection, provenance tracking, and user-centric evaluations to promote more transparent and equitable real-world applications.

\section*{Acknowledgments}
This work was supported by the Institute for Information and communications Technology Promotion (IITP) grant funded by the Korea government (MSIT) (No. 2022-0-00010, Development of Korean sign language translation service technology for the deaf in medical environment).

\bibliography{custom}
\clearpage

\appendix



\section*{Appendix}

In the Appendix, we provide additional implementation details and present supplementary results and analyses not covered in the main text.

\section{More Related Work}
\label{app:related_work}
\subsection{Information Retrieval}
Advancements in information retrieval (IR) have improved retrieval efficiency, granularity, and ranking quality. Encoder-based models, such as DPR~\cite{DPR-karpukhin} and Contriever~\cite{contriever}, have been widely adopted for retrieval tasks. More recently, M3-Embedding~\cite{bge-m3} unifies dense, multi-vector, and sparse retrieval via self-knowledge distillation, enabling retrieval across different text granularities.

Decoder-only LLMs have emerged as strong rerankers, leveraging large-scale data for improved ranking at the cost of computational complexity. LLaRA~\cite{llara} and RePLlama~\cite{repllama} employ dense retrieval and multi-stage ranking, while UPR~\cite{upr} and ranking prompting methods~\cite{holistic, prp} refine retrieval through generation-based approaches. However, these methods focus on reordering passages rather than compressing retrieved content.

Research on retrieval granularity~\cite{phrase, phrase_retriever, proconvqa, denseXretrieval, dslr} explores balancing precision and contextual coherence. While fine-grained retrieval enhances specificity, excessive fragmentation can distort meaning~\cite{decontextualization}. Retrieval and ranking aim to maximize recall but do not address post-retrieval redundancy. EXIT operates beyond retrieval and ranking, focusing on adaptive post-retrieval context compression, which selectively filters content to optimize both efficiency and relevance in RAG.

\subsection{Summarization}
Summarization methods are broadly categorized as extractive and abstractive. Extractive summarization retains factual consistency by selecting key sentences but often suffers from redundancy and coherence issues~\cite{extractive, extractive2}. Abstractive summarization improves fluency and readability but is computationally intensive and prone to hallucination~\cite{abstractive, abstractive2}. Despite these challenges, its ability to generate concise outputs has driven interest in its application~\cite{feedbacksummary, unisumeval, FineSurE}.

Query-focused summarization refines document content based on relevance to a given query, removing unrelated information while preserving key details~\cite{qfs, qfs2}. However, traditional summarization methods process single documents and prioritize completeness and faithfulness~\cite{FineSurE, feedbacksummary}. By contrast, post-retrieval compression in RAG requires query-adaptive filtering across multiple retrieved documents. EXIT fills this gap by providing a dynamic and query-adaptive compression strategy, distinguishing between relevant, marginal, and irrelevant content while maintaining both efficiency over latency and answer accuracy.

\section{More Implementation Details}
\label{app:implementation}

This section describes our training environment, data composition, and prompt templates. All experiments were conducted on an NVIDIA A100-SXM4-80GB GPU cluster. Training was performed on a single GPU with gradient accumulation.

\subsection{Training Configuration}
We employed LoRA~\cite{lora} to fine-tune the compressor model, enabling parameter-efficient training while preserving performance. The model was trained with the following hyperparameters:  
\begin{itemize}[nosep,leftmargin=*]  
    \item Batch size: 8 per device  
    \item Gradient accumulation steps: 8  
    \item Learning rate: 1e-5  
    \item Weight decay: 0.1  
    \item Warmup ratio: 0.03  
    \item Training epochs: 1  
    \item Optimizer: \texttt{paged\_adamw\_8bit}  
    \item Quantization: 4-bit with float16 precision  
    \item LoRA configuration: Rank = 64, Scaling = 32, Dropout = 0.05  
\end{itemize}  

\subsection{Model Selection and Training Time}
Model selection was guided by validation loss. Training was conducted on our cluster, requiring approximately 90 hours.

\subsection{Data Processing}

\begin{table}[h!]
\centering
\caption{Statistics of the training dataset constructed from HQA. Positive (Pos) sentences are required for the correct answer, Hard-Neg (H-Neg) sentences appear in the same passages but lack crucial evidence, and Neg sentences come from unrelated queries. Counts are in thousands (K).}
\label{tab:trainingset}
\begin{tabular}{l|rrrr}
\toprule[1.5pt]
Split & Pos & H-Neg & Neg & Total \\
\midrule[1pt]
Train & 213K & 107K & 107K & 427K \\
Valid & 2.4K & 1.2K & 1.2K & 4.8K \\
\bottomrule[1.5pt]
\end{tabular}
\end{table}

We used SpaCy to segment documents into sentences. Table~\ref{tab:trainingset} shows the composition of the training and validation sets derived from HQA. The training set contains 427K sentences, including 213K positive (Pos), 107K hard-negative (H-Neg), and 107K negative (Neg) instances. The validation set includes 4.8K sentences with a similar distribution. This balanced composition ensures the classifier encounters diverse retrieval scenarios during training.

\subsection{Inference Settings}
For inference, we set:
\begin{itemize}[nosep,leftmargin=*]
    \item Temperature: 0.0
    \item Top-p: 1.0
    \item vLLM version: v0.5.5
    \item Relevance threshold $(\tau)$: 0.5
\end{itemize}

\subsection{Prompt Templates}
\begin{table}[h!]
\centering
\caption[A prompt template for document compression.]{A prompt template for document compression.}
\begin{tcolorbox}[width=\columnwidth, colback=white, title={Compression Prompt Template}, colbacktitle=white, coltitle=black]
Query:

\{query\}

Full context:

\{original passage\}

Sentence:

\{sentence\}

Is this sentence useful in answering the query? Answer only ``Yes'' or ``No''.
\end{tcolorbox}
\label{compression_prompt_tab}
\end{table}

\begin{table}[h!]
\caption[A prompt template used by the reader model for the QA task.]{A prompt template used by the reader model for the QA task.} 
\begin{tcolorbox}
[width=\columnwidth,colback={white},title={QA Prompt Template},colbacktitle=white,coltitle=black]

Context information is below.

---------------------

\{context\}

---------------------

Given the context information and not prior knowledge, answer the query. Do not provide any explanation.

Query: \{query\}

Answer: 
\end{tcolorbox}
\label{qa_prompt_tab}
\end{table}

Full prompt templates for compression and QA tasks are provided in Tables~\ref{compression_prompt_tab} and \ref{qa_prompt_tab}, respectively.

\section{Additional Experimental Results and Analyses}
\label{app:Experimental}

Table~\ref{tab:detailed_performance_comparison} shows detailed results for each dataset and model configuration under zero-shot QA prompts, using both Top-5 and Top-20 retrieval. For the few-shot QA prompts, Table~\ref{tab:performance_comparison_fewshot} summarizes the results, where we randomly selected five training examples per dataset as demonstrations. Table~\ref{tab:token_statistics} provides comprehensive token count statistics, and Table~\ref{tab:compression_time_comparison} breaks down end-to-end latency.

\subsection{Performance with Sparse Retrieval}
\begin{table}[]
    \centering
    \small
    \caption[Performance Comparison with Sparse Retrieval]{Performance comparison between in-domain (HQA) and out-of-domain (2WIKI) datasets using BM25 as the retriever model. Best results are highlighted in \textbf{bold}, and second best results are \underline{underlined}.}
    \label{tab:sparse_retrieval}
    \setlength{\tabcolsep}{3pt}
    \renewcommand{\arraystretch}{1.1}
    \small
    \begin{adjustbox}{width=\columnwidth}
    \begin{tabular}{l|c|ccc|ccc}
        \toprule[1.5pt]
        \multirow{2}{*}{\textbf{Compressor}} &\multirow{2}{*}{\textbf{Type}} 
        & \multicolumn{3}{c|}{\textbf{HQA}}
        & \multicolumn{3}{c}{\textbf{2WIKI}} \\
        \cmidrule[0.8pt]{3-8}
        &
        & \scriptsize EM $\uparrow$ & \scriptsize F1 $\uparrow$ & \scriptsize \#token (\%) $\downarrow$ & \scriptsize EM $\uparrow$ & \scriptsize F1 $\uparrow$ & \scriptsize \#token (\%) $\downarrow$ \\
        \midrule[1pt]
        \multicolumn{8}{c}{\textit{Top-5 Documents}} \\
        \midrule[0.5pt]
        Original Docs & -
        & 28.2 & 39.1 & 755.8 (100.0) & 19.6 & 25.9 & 789.7 (100.0) \\
        RECOMP-Abst & Abs.
        & 27.8 & 39.2 & \textbf{63.0 (8.3)} & \textbf{25.0} & \textbf{30.6} & \textbf{55.7 (7.1)} \\
        CompAct & Abs.
        & 30.6 & 41.2 & \underline{76.9 (10.2)} & 19.6 & \underline{29.1} & 71.0 (9.0) \\
        Refiner & Abs.
        & \underline{30.8} & \underline{42.3} & 84.0 (11.1) & 19.6 & 27.8 & \underline{62.9 (8.0)} \\
        RECOMP-Extr & Ext.
        & 28.2 & 37.5 & 93.1 (12.3) & 11.8 & 19.1 & 99.1 (12.5)\\
        LongLLMLingua & Ext.
        & 28.6 & 39.7 & 230.3 (30.5) & 22.2 & 27.4 & 236.6 (30.0) \\
        \rowcolor{blue!5} EXIT (Ours) & Ext.
        & \textbf{33.4} & \textbf{44.5} & 177.8 (23.5) & \underline{24.4} & \underline{29.1} & 138.2 (17.5)\\
        \midrule[0.5pt]
        \multicolumn{8}{c}{\textit{Top-20 Documents}} \\
        \midrule[0.5pt]
        Original Docs & -
        & 31.2 & 42.5 & 3009.5 (100.0) & 23.0 & \underline{30.6} & 3132.5 (100.0) \\
        RECOMP-Abst & Abs.
        & 29.0 & 39.7 & \textbf{69.7 (2.3)} & 22.8 & 28.2 & \textbf{52.3 (1.7)} \\
        CompAct & Abs.
        & \underline{31.6} & \underline{43.0} & 109.5 (3.6) & 20.4 & 28.7 & 113.2 (3.6)\\
        Refiner & Abs.
        & 28.4 & 38.8 & 136.1 (4.5) & 18.8 & 26.9 & 108.4 (3.5)\\
        RECOMP-Extr & Ext.
        & 28.4 & 37.0 & \underline{93.5 (3.1)} & 11.2 & 19.3 & \underline{96.7 (3.1)} \\
        LongLLMLingua & Ext.
        & 31.0 & 40.7 & 558.0 (18.5)& \underline{23.6} & 28.3 & 593.7 (19.0) \\
        \rowcolor{blue!5} EXIT (Ours) & Ext.
        & \textbf{35.2} & \textbf{46.9} & 411.3 (13.7) & \textbf{27.2} & \textbf{32.4} & 312.7 (10.0)\\
        \bottomrule[1.5pt]
    \end{tabular}
    \end{adjustbox}
\end{table}

To assess EXIT’s robustness with different retrieval architectures, we evaluate it with BM25, a sparse retrieval method. Table~\ref{tab:sparse_retrieval} compares EXIT’s performance on HQA (in-domain) and 2WIKI (out-of-domain) under Top-5 and Top-20 retrieval settings.

With Top-5 retrieval, EXIT shows notable gains on HQA, improving EM (33.4 vs. 28.2) and F1 (44.5 vs. 39.1) over the uncompressed baseline. Although RECOMP-Abst performs best on 2WIKI (25.0 EM, 30.6 F1), EXIT remains competitive (24.4 EM, 29.1 F1).

EXIT’s advantages grow with Top-20 retrieval. On HQA, EXIT outperforms all baselines, improving EM by 4.0 points (35.2 vs. 31.2) and F1 by 4.4 points (46.9 vs. 42.5) compared to using uncompressed documents. On 2WIKI, EXIT achieves the highest scores (27.2 EM, 32.4 F1), confirming its generalizability across domains and retrieval strategies.




\subsection{Performance with Proprietary Model}
\begin{table}[]
   \centering
   \small
   \caption[Performance Comparison with Closed-Source Proprietary Model]{Performance comparison between in-domain (HQA) and out-of-domain (2WIKI) datasets using GPT-4o as the reader model. Best results are highlighted in \textbf{bold}, and second best results are \underline{underlined}.}
   \label{tab:gpt_result}
   \setlength{\tabcolsep}{3pt}
   \renewcommand{\arraystretch}{1.1}
   \small
   \begin{adjustbox}{width=\columnwidth}
   \begin{tabular}{l|c|ccc|ccc}
        \toprule[1.5pt]
        \multirow{2}{*}{\textbf{Compressor}} & \multirow{2}{*}{\textbf{Type}}
        & \multicolumn{3}{c|}{\textbf{HQA}}
        & \multicolumn{3}{c}{\textbf{2WIKI}} \\
        \cmidrule[0.8pt]{3-8}
        &
        & \scriptsize EM $\uparrow$ & \scriptsize F1 $\uparrow$ & \scriptsize \#token (\%) $\downarrow$ & \scriptsize EM $\uparrow$ & \scriptsize F1 $\uparrow$ & \scriptsize \#token (\%) $\downarrow$ \\
        \midrule[1pt]
        \multicolumn{8}{c}{\textit{Top-5 Documents}} \\
        \midrule[0.5pt]
        Original Docs & -
        & 37.2 & \underline{48.6} & 735.3 (100.0) & \underline{31.2} & \underline{35.3} & 764.5 (100.0) \\
        RECOMP-Abst & Abs.
        & 29.4 & 40.2 & \textbf{62.8 (8.5)} & 23.8 & 27.8 & \textbf{53.5 (7.0)} \\
        CompAct & Abs.
        & \underline{37.4} & 48.0 & 74.3 (10.1) & 30.0 & 33.6 & 67.6 (8.8) \\
        Refiner & Abs.
        & 35.8 & 47.5 & \underline{71.4 (9.7)} & 27.8 & 32.6 & \underline{54.2 (7.1)} \\
        RECOMP-Extr & Ext.
        & 32.4 & 41.7 & 87.1 (11.8) & 25.6 & 28.6 & 93.6 (12.2) \\
        LongLLMLingua & Ext.
        & 34.2 & 45.2 & 223.1 (30.3) & 30.6 & 34.5 & 230.5 (30.2) \\
        \rowcolor{blue!5} EXIT (Ours) & Ext.
        & \textbf{38.2} & \textbf{50.4} & 191.2 (26.0) & \textbf{31.8} & \textbf{35.8} & 145.6 (19.0) \\
        \midrule[0.8pt]
        \multicolumn{8}{c}{\textit{Top-20 Documents}} \\
        \midrule[0.5pt]
        Original Docs & -
        & \textbf{39.6} & \textbf{51.8} & 2940.5 (100.0) & \textbf{40.0} & \textbf{43.8} & 3066.2 (100.0) \\
        RECOMP-Abst & Abs.
        & 33.6 & 44.2 & \textbf{62.7 (2.1)} & 26.6 & 32.1 & \textbf{48.9 (1.6)} \\
        CompAct & Abs.
        & 33.0 & 43.7 & 106.0 (3.6) & 23.0 & 27.3 & 105.1 (3.4) \\
        Refiner & Abs.
        & 31.8 & 41.5 & 130.6 (4.4) & 31.0 & 35.5 & 100.3 (3.3)\\
        RECOMP-Extr & Ext.
        & 31.2 & 39.6 & \underline{86.2 (2.9)} & 23.2 & 27.2 & \underline{91.0 (3.0)} \\
        LongLLMLingua & Ext.
        & 38.8 & 49.4 & 549.5 (18.7) & 35.4 & 39.6 & 581.5 (19.0) \\
        \rowcolor{blue!5} EXIT (Ours) & Ext.
        & \underline{39.4} & \underline{50.1} & 453.6 (15.4) & \underline{35.6} & \underline{40.3} & 346.7 (11.3) \\
        \bottomrule[1.5pt]
    \end{tabular}
    \end{adjustbox}
\end{table}

We further examine EXIT’s effectiveness using GPT-4o as the reader. Table~\ref{tab:gpt_result} compares performance on HQA (in-domain) and 2WIKI (out-of-domain), along with compression rates.

For Top-5 retrieval, EXIT attains the best accuracy on HQA (38.2 EM, 50.4 F1) while retaining only 26.0\% of tokens. This surpasses the uncompressed baseline (37.2 EM, 48.6 F1) with a 74\% token reduction. On 2WIKI, EXIT maintains leading accuracy (31.8 EM, 35.8 F1) while using just 19.0\% of the original tokens.

Under Top-20 retrieval, where uncompressed documents benefit from greater coverage, EXIT still achieves competitive accuracy with substantially fewer tokens. On HQA, EXIT closely matches the uncompressed EM score (39.4 vs. 39.6) while using only 15.4\% of tokens. Although RECOMP variants compress more aggressively, they suffer marked performance drops. LongLLMLingua performs similarly to EXIT but retains more tokens (18.7\% vs. 15.4\%).

These findings illustrate EXIT’s ability to balance performance and efficiency, making it valuable for API-based proprietary models where token costs and accuracy both matter.

\subsection{Evaluation in Specialized Domains}
\begin{table}[t]
\small
\centering
\caption{Evaluation of EXIT and baselines in specialized domains. BioASQ evaluates biomedical QA using PubMed documents, and COVID-QA focuses on COVID-specific QA over the TREC-COVID corpus.}
\label{tab:domain-results}
\begin{tabular}{lccc}
\toprule
\multicolumn{4}{c}{\textbf{BioASQ (Biomedical QA)}} \\
\midrule
\textbf{Method} & \textbf{EM} & \textbf{F1} & \textbf{Avg. \#Tokens} \\
\midrule
Original Docs     & 57.2 & 68.6 & 1791.49 \\
RECOMP-Abst       & 54.6 & 63.8 & 37.72 \\
CompAct           & 58.0 & 68.2 & 94.81 \\
Refiner           & 58.8 & 68.5 & 229.22 \\
RECOMP-Extr       & 55.8 & 65.0 & 109.72 \\
LongLLMLingua     & 58.2 & 66.4 & 364.06 \\
\rowcolor{blue!5} \textbf{EXIT (Ours)} & \textbf{58.6} & \textbf{67.1} & \textbf{132.28} \\
\midrule
\multicolumn{4}{c}{\textbf{COVID-QA (COVID-19 QA)}} \\
\midrule
\textbf{Method} & \textbf{EM} & \textbf{F1} & \textbf{Avg. \#Tokens} \\
\midrule
Original Docs     & 3.8  & 16.0 & 1775.66 \\
RECOMP-Abst       & 2.2  & 11.5 & 17.94 \\
CompAct           & 3.4  & 14.6 & 95.65 \\
Refiner           & 3.8  & 14.6 & 201.38 \\
RECOMP-Extr       & 1.4  & 11.4 & 103.11 \\
LongLLMLingua     & 2.2  & 13.5 & 353.03 \\
\rowcolor{blue!5} \textbf{EXIT (Ours)} & \textbf{5.8}  & \textbf{17.2} & \textbf{129.33} \\
\bottomrule
\end{tabular}
\end{table}

To assess the generalizability of EXIT beyond general-domain datasets, we conducted additional evaluations on two specialized QA benchmarks: 1) \textbf{BioASQ} \cite{bioasq}, a biomedical QA dataset using the PubMed corpus (14.9M documents), 2)\textbf{ COVID-QA} \cite{covidqa}, a COVID-specific QA benchmark based on the TREC-COVID collection \cite{treccovid} (171K documents).

Although our classifier was trained solely on HQA, a general-domain dataset, EXIT performs competitively or better than existing baselines in both specialized domains. Table~\ref{tab:domain-results} shows that EXIT achieves high EM and F1 scores while substantially reducing token counts. On COVID-QA, EXIT notably outperforms all baselines in EM and F1 despite the domain shift.

These results suggest that EXIT’s context-aware extraction strategy generalizes well, preserving key evidence without requiring domain-specific fine-tuning. This highlights EXIT’s robustness across diverse retrieval scenarios.

\subsection{Evaluation on Complex Multi-hop Reasoning}
\begin{table}[t]
\small
\centering
\caption{Evaluation of EXIT and baselines on the MuSiQue dataset on Llama3-8B reader.}
\label{tab:musique-results}
\begin{tabular}{lccc}
\toprule
\textbf{Method} & \textbf{EM} & \textbf{F1} & \textbf{Latency (s)} \\
\midrule
Original Docs     & 3.8  & 10.3 & 1.1 \\
RECOMP-Abst       & 4.5  & 11.8 & 2.5 \\
CompAct           & 4.1  & 11.0 & 8.5 \\
Refiner           & 4.5  & 11.3 & 7.6 \\
RECOMP-Extr       & 4.3  & 10.1 & \textbf{0.6} \\
LongLLMLingua     & 4.0  & 10.7 & 0.9 \\
\rowcolor{blue!5} \textbf{EXIT (Ours)} & \textbf{4.6}  & \textbf{11.3} & 0.9 \\
\bottomrule
\end{tabular}
\end{table}

To further assess EXIT’s ability to handle complex reasoning tasks, we evaluated it on the MuSiQue dataset \cite{musique}, which contains multi-hop questions requiring deeper reasoning (3–4 hops). While our main experiments focused on standard benchmarks like HQA and 2WIKI, MuSiQue provides a valuable test of generalization under more challenging conditions.

As shown in Table~\ref{tab:musique-results}, EXIT achieves competitive performance on MuSiQue. Notably, it maintains low latency while matching or outperforming other extractive and abstractive baselines in EM and F1 scores. These results demonstrate EXIT’s robustness in handling longer reasoning chains without incurring the high latency costs typically associated with abstractive methods.

These findings reinforce EXIT’s ability to scale to more complex, multi-hop QA settings while maintaining an efficient trade-off between latency and performance.

\subsection{Latency Measurement and Analysis}
\begin{table*}[t]
\small
\centering
\caption{Average end-to-end latency (seconds per sample) across different compression methods. Results are the mean and standard deviation (Std) over five runs. Experiments for the 8B reader were conducted on a single A100-80GB SXM GPU, and for the 70B reader on four A100-80GB SXM GPUs.}
\label{tab:fivetimeslatency}
\begin{tabular}{lcccccc}
\toprule
\textbf{8B Reader} & \multicolumn{2}{c}{\textbf{Compression}} & \multicolumn{2}{c}{\textbf{Reading}} & \multicolumn{2}{c}{\textbf{Total}} \\
\cmidrule(r){2-3} \cmidrule(r){4-5} \cmidrule(r){6-7}
\textbf{Method} & Avg & Std & Avg & Std & Avg & Std \\
\midrule
Original Docs     & --      & --      & 0.9481 & 0.0056 & 0.9481 & 0.0056 \\
RECOMP-Abst       & 1.9967  & 0.0159  & 0.3177 & 0.0008 & 2.3144 & 0.0167 \\
CompAct           & 12.4682 & 0.0030  & 0.3930 & 0.0006 & 12.8612 & 0.0036 \\
Refiner           & 6.3881  & 0.0332  & 0.3877 & 0.0012 & 6.7758 & 0.0344 \\
RECOMP-Extr       & 0.0279  & 0.0022  & 0.3668 & 0.0040 & 0.3947 & 0.0062 \\
LongLLMLingua     & 0.3721  & 0.0050  & 0.4526 & 0.0004 & 0.8247 & 0.0054 \\
\rowcolor{blue!5} \textbf{EXIT (Ours)} & \textbf{0.4263} & \textbf{0.0031} & \textbf{0.4529} & \textbf{0.0028} & \textbf{0.8792} & \textbf{0.0059} \\
\midrule
\textbf{70B Reader} & \multicolumn{2}{c}{\textbf{Compression}} & \multicolumn{2}{c}{\textbf{Reading}} & \multicolumn{2}{c}{\textbf{Total}} \\
\cmidrule(r){2-3} \cmidrule(r){4-5} \cmidrule(r){6-7}
\textbf{Method} & Avg & Std & Avg & Std & Avg & Std \\
\midrule
Original Docs     & --      & --      & 8.0835 & 0.0072 & 8.0835 & 0.0072 \\
RECOMP-Abst       & 1.8621  & 0.0171  & 2.3210 & 0.0113 & 4.1831 & 0.0284 \\
CompAct           & 15.1999 & 0.0132  & 2.6932 & 0.0007 & 17.8931 & 0.0139 \\
Refiner           & 8.3520  & 0.0248  & 3.0993 & 0.0017 & 11.4513 & 0.0265 \\
RECOMP-Extr       & 0.0329  & 0.0069  & 2.8288 & 0.0031 & 2.8617 & 0.0100 \\
LongLLMLingua     & 0.4788  & 0.0127  & 3.7121 & 0.0012 & 4.1909 & 0.0139 \\
\rowcolor{blue!5} \textbf{EXIT (Ours)} & \textbf{0.3253} & \textbf{0.0086} & \textbf{3.5341} & \textbf{0.0022} & \textbf{3.8594} & \textbf{0.0108} \\
\bottomrule
\end{tabular}
\end{table*}

\noindent
To ensure the reliability and reproducibility of our latency results, we conducted five independent runs for each compression method and report both the mean and standard deviation of end-to-end latency (measured in seconds per sample). End-to-end latency includes both compression time and reading/inference time, measured from the moment a request is issued to the moment the answer is returned by the model.

For consistency, all experiments using the 8B reader were run on a single A100-80GB SXM GPU, while those using the 70B reader were executed on four A100-80GB SXM GPUs. To minimize hardware-induced noise, we ensured that no unrelated processes were active during measurement and maintained fixed hyperparameters and model configurations across all runs.

As shown in Table~\ref{tab:fivetimeslatency}, our method EXIT achieves a strong balance between latency and accuracy. While abstractive methods such as CompAct and Refiner incur significant latency (12.86s and 6.78s ,respectively, with the 8B reader), EXIT completes inference in under 1 second (0.88s) on average. Even with the 70B reader, EXIT remains efficient with 3.86s total latency, significantly faster than the original uncompressed baseline (8.08s), while also improving QA performance.

These results reinforce EXIT’s practicality for real-world deployment, particularly in latency-sensitive applications.

\subsection{Impact of Threshold $\tau$}
\begin{figure*}[t]
\centerline{\includegraphics[width=\textwidth]{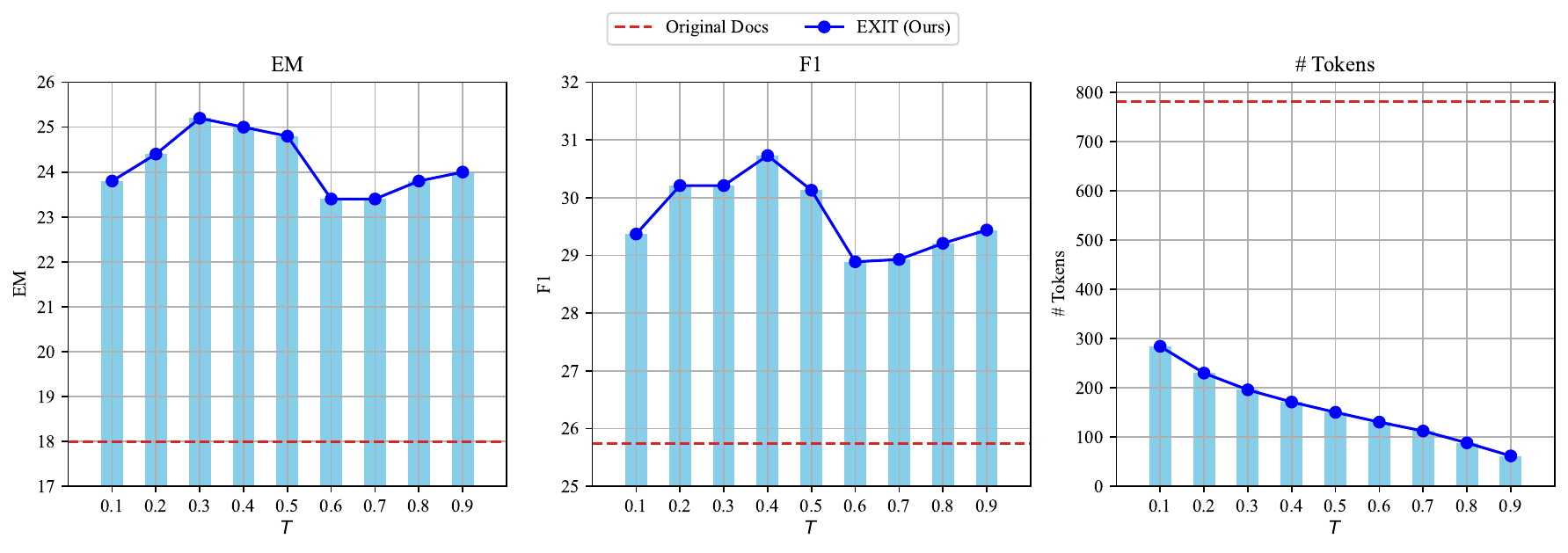}}
    \caption[Impact of Threshold $\tau$ on Retained Sentences]{Changes in EM, F1 score, and token count as the threshold $\tau$ for retaining sentences is adjusted.} \label{fig:threshold}
\end{figure*}
\noindent
We analyze EXIT’s sensitivity to the relevance threshold \(\tau\). Figure~\ref{fig:threshold} shows EXIT’s performance across various \(\tau\) values.

EXIT remains stable over a wide threshold range, with strong results between \(\tau\)=0.3–0.5. At \(\tau\)=0.3, EXIT reaches 25.2 EM using only 25\% of the tokens (195.82 vs. 780.95 for the baseline), a substantial improvement over the original documents (18.0 EM). F1 scores also remain consistently higher than the baseline (30.21–30.73 vs. 25.74).

Even under extreme compression (\(\tau\)=0.9, 7.9\% of tokens), EXIT achieves better accuracy (24.0 EM, 29.44 F1) than the uncompressed documents. Conversely, a lenient threshold (\(\tau\)=0.1) retains more tokens but still provides benefits, demonstrating that EXIT effectively identifies crucial content under varying conditions.

This robustness across thresholds gives practitioners flexibility to adjust the compression-accuracy trade-off without severely impacting performance.

\subsection{Effect of Classifier Training}
\begin{table}[t]
\small
\centering
\caption{Comparison of EXIT using a zero-shot (frozen Gemma-2B) vs. fine-tuned classifier on HotpotQA (HQA) and 2WikiMultiHopQA (2WIKI).}
\label{tab:classifier-training}
\begin{tabular}{lccc}
\toprule
\textbf{Dataset} & \textbf{Method} & \textbf{EM} & \textbf{F1} \\
\midrule
\multirow{3}{*}{2WIKI} 
& Original Docs      & 18.0 & 25.7 \\
& EXIT (Zero-shot)   & 22.4 & 27.9 \\
& EXIT (Ours)        & \textbf{24.8} & \textbf{30.1} \\
\midrule
\multirow{3}{*}{HQA} 
& Original Docs      & 29.2 & 40.2 \\
& EXIT (Zero-shot)   & 31.4 & \textbf{43.1} \\
& EXIT (Ours)        & \textbf{31.6} & 42.6 \\
\bottomrule
\end{tabular}

\vspace{0.5em}

\begin{tabular}{lcc}
\toprule
\textbf{Dataset} & \textbf{Method} & \textbf{Avg. \#Tokens} \\
\midrule
\multirow{3}{*}{2WIKI} 
& Original Docs    & 764.47 \\
      & EXIT (Zero-shot) & 731.35 \\
      & EXIT (Ours)      & \textbf{145.62} \\
\midrule
\multirow{3}{*}{HQA}
& Original Docs    & 735.24 \\
      & EXIT (Zero-shot) & 727.87 \\
      & EXIT (Ours)      & \textbf{191.11} \\
\bottomrule
\end{tabular}
\end{table}

To evaluate the impact of fine-tuning the classifier in EXIT, we compared our trained model with a frozen zero-shot version using Gemma-2B (without any task-specific fine-tuning). We tested both variants on an in-domain dataset (HQA) and an out-of-domain dataset (2WIKI).

As shown in Table~\ref{tab:classifier-training}, the zero-shot classifier already improves over the uncompressed baseline, indicating the strength of relevance-based filtering even without training. However, our fine-tuned classifier consistently outperforms the zero-shot variant in both EM and F1 scores while achieving significantly greater token reduction—demonstrating that supervised training yields both better accuracy and more efficient compression.

\subsection{Adaptability of EXIT to Different Supervision Signals}
\label{app:signal}
\begin{table}[t]
\centering
\caption{Comparison of EXIT with baselines on QA tasks on HQA. EXIT (GPT-4o) refers to a variant using GPT-4o as a zero-shot relevance classifier, while EXIT (Ours) represents our proposed method trained with supervised data.}
\small
\begin{tabular}{lccc}
\toprule
\textbf{Method} & \textbf{EM} & \textbf{F1} & \textbf{\#token} \\
\midrule
Original Docs & 29.2 & 40.2 & 735.2 \\
RECOMP-Extr & 25.2 & 34.8 & 89.7 \\
RECOMP-Abst & 19.8 & 24.7 & 55.5 \\
LongLLMLingua & 27.4 & 40.2 & 227.1 \\
CompAct & 29.4 & 40.9 & 76.3 \\
Refiner & 28.8 & 40.7 & 73.4 \\
\midrule
EXIT (GPT-4o) & 30.4 & 41.9 & 84.2 \\
EXIT (Ours) & \textbf{31.6} & \textbf{42.6} & 191.0 \\
\bottomrule
\end{tabular}
\label{tab:exit_vs_baselines}
\end{table}

A key strength of EXIT is its ability to integrate different classifiers for sentence selection without being constrained to manually annotated datasets. To explore this flexibility, we evaluate an alternative approach where relevance scores are derived from GPT-4o without explicit fine-tuning. As shown in Table~\ref{tab:exit_vs_baselines}, EXIT maintains strong performance, outperforming or closely matching baselines such as LongLLMLingua and CompAct. This demonstrates that EXIT can leverage diverse supervision signals while remaining adaptable to different scoring mechanisms. Moreover, these results suggest the potential of utilizing large-scale pseudo-labeled data to further refine EXIT's training, enhancing scalability without relying strictly on human-labeled datasets. This adaptability highlights EXIT’s robustness and practical applicability across various retrieval settings.

\subsection{Analysis of Classification Performance Across Negative Sample Types}


\begin{table}[]
    \centering
    \caption{Classification performance for Yes/No labels.}
    \label{tab:table5.2}
    \setlength{\tabcolsep}{4.5pt}
    \renewcommand{\arraystretch}{1.1}
    \begin{adjustbox}{width=0.3\textwidth}
    \begin{tabular}{l|ccc}
        \toprule[1.5pt]
        \multicolumn{4}{c}{\cellcolor{gray!10}\textbf{Overall}} \\
        \midrule[0.8pt]
        \multirow{1}{*}{\textbf{Class}} 
        & \textbf{Precision} $\uparrow$
        & \textbf{Recall} $\uparrow$
        & \textbf{F1-Score} $\uparrow$ \\
        \midrule[0.8pt]
        Yes & 0.91 & \textbf{0.93} & \textbf{0.92} \\
        No & \textbf{0.93} & 0.91 & \textbf{0.92} \\
        \midrule[1pt]
        \multicolumn{4}{c}{\cellcolor{gray!10}\textbf{Hard Negative}} \\
        \midrule[0.8pt]
        Yes & 0.86 & \textbf{0.93} & \textbf{0.89} \\
        No & \textbf{0.93} & 0.84 & 0.88 \\
        \midrule[1pt]
        \multicolumn{4}{c}{\cellcolor{gray!10}\textbf{Negative}} \\
        \midrule[0.8pt]
        Yes & \textbf{0.96} & 0.93 & \textbf{0.95} \\
        No & 0.93 & \textbf{0.96} & \textbf{0.95} \\
        \bottomrule[1.5pt]
    \end{tabular}
    \end{adjustbox}
\end{table}

Table~\ref{tab:table5.2} presents EXIT’s sentence-level classification performance, broken down by negative sample type. EXIT achieves 0.92 F1 for both positive (“Yes”) and negative (“No”) classes overall, indicating a balanced ability to identify essential and non-essential sentences.

The model excels at filtering random negatives (0.95 F1), effectively discarding irrelevant content. With hard negatives (topically related but not essential), EXIT still performs well (0.89 F1 for positive, 0.88 for negative), handling nuanced relevance distinctions.

These results highlight EXIT’s adaptability and confirm its suitability for real-world RAG scenarios where both overtly irrelevant and subtly extraneous content must be managed.
\begin{figure}[t]
\centerline{\includegraphics[width=\linewidth]{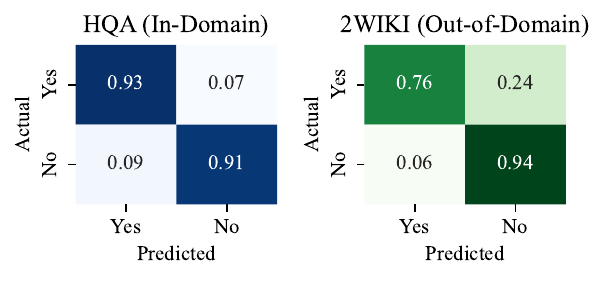}}
    \caption{Confusion matrices (row-normalized) for context-aware relevance classification on HQA (in-domain) and 2WIKI (out-of-domain).} \label{fig:confusion_matrix}
\end{figure}
\subsection{Classification Performance}
To better understand the effectiveness of our context-aware relevance classifier, we report row-normalized confusion matrices for both in-domain (HQA) and out-of-domain (2WIKI) datasets, as shown in Figure~\ref{fig:confusion_matrix}. On HQA, the classifier displays a perfectly balanced ability to recognize both relevant (“Yes”) and irrelevant (“No”) sentences, achieving over 90\% precision and recall in each category. 
While, on the 2WIKI dataset, the classifier exhibits a slight drop in recall for ``Yes'' sentences, it still shows a strong classification ability with over 70\% recall and 90\% precision.

These results confirm that the classifier performs robustly in its training domain and generalizes reasonably well to unseen queries, though we leave narrowing this discrepancy as a valuable future research direction.

\subsection{Classification Performance under Ablation Settings}
\label{app:classification_ablation}
\subsubsection{Analysis of Training Data Composition}

\begin{figure}[t]
\centerline{\includegraphics[width=\linewidth]{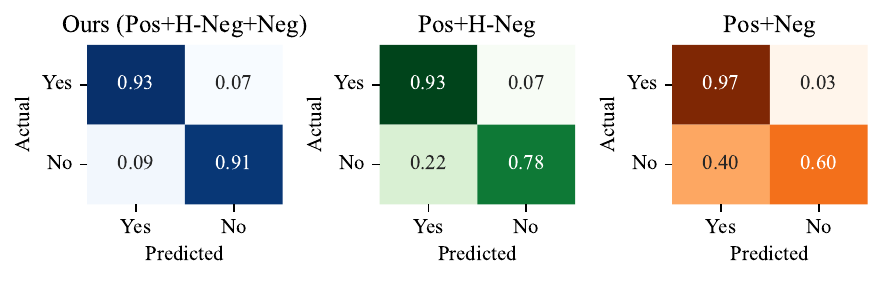}}
    \vspace{-0.15in}
    \caption{Row-normalized confusion matrices for classification performance under different training data conditions: Ours (Pos+H-Neg+Neg), Pos+H-Neg, and Pos+Neg. Each matrix compares the predicted (“Yes”/“No”) labels against the actual labels.} \label{fig:confusion_matrix_ablation}
\end{figure}

Figure~\ref{fig:confusion_matrix_ablation} presents row-normalized confusion matrices comparing classification performance across three training data configurations: Ours (Pos+H-Neg+Neg), Pos+H-Neg, and Pos+Neg. Under the Ours setup, the classifier displays a balanced ability to identify both “Yes” (relevant) and “No” (irrelevant) sentences, achieving an F1-score of 0.92 for both classes. By contrast, excluding one type of negative sample (Pos+H-Neg or Pos+Neg) reduces overall robustness, evidenced by declines in both accuracy and class-wise F1 scores. For instance, the Pos+Neg configuration struggles to maintain balance, accurately identifying “Yes” instances but misclassifying a substantial number of “No” cases. These results confirm that incorporating a comprehensive mix of positive, hard-negative, and random-negative samples leads to more reliable and contextually aware sentence selection, thereby improving the classifier’s performance in practical retrieval-augmented QA scenarios.

\subsubsection{Impact of Context on Classification Performance}
\begin{figure}[t]
\centerline{\includegraphics[width=\linewidth]{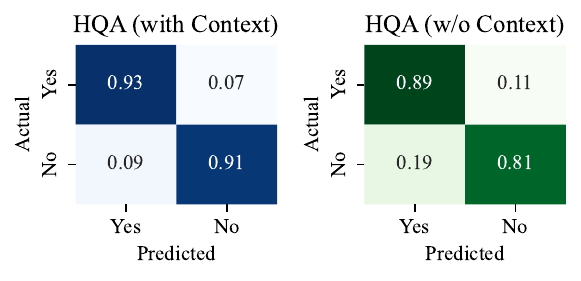}}
    \vspace{-0.1in}
    \caption{Row-normalized confusion matrices comparing classification performance with (left) and without (right) contextual information on HQA. The availability of context improves the model’s ability to accurately distinguish relevant (“Yes”) from non-relevant (“No”) sentences.} \label{fig:confusion_matrix_context}
\end{figure}

We evaluate the classifier’s performance with and without broader passage-level context. Figure~\ref{fig:confusion_matrix_context} shows that including context maintains over 90\% precision and recall for both “Yes” and “No” classes. Without context, precision and recall decline, weakening the distinction between relevant and irrelevant sentences. This emphasizes the importance of incorporating passage-level context for accurately identifying answer-critical information.




\subsection{Sentence Reordering Strategy Analysis}
\begin{table}[t]
\centering
\caption{Ablation results comparing sentence reordering strategies in EXIT on HotpotQA. ``Random'' shuffles selected sentences, while ``Ascending'' and ``Descending'' sort by relevance scores. ``Original order'' maintains the sentence sequence from the original document.}
\label{tab:reorder}
\begin{tabular}{lcc}
\toprule
\textbf{Reordering Strategy} & \textbf{EM} & \textbf{F1} \\
\midrule
Random                    & 30.20 & 41.30 \\
Ascending (score-based)   & 30.43 & 41.44 \\
Descending (score-based)  & 30.33 & 41.16 \\
\textbf{Original order (EXIT)}     & \textbf{30.61} & \textbf{41.49} \\
\bottomrule
\end{tabular}
\end{table}

One potential concern with extractive compression is that removing many sentences may lead to unnatural or ambiguous contexts when the remaining content is reassembled. In EXIT, we preserve the original sentence order during reconstruction to maintain document coherence, following prior work \cite{dslr}. However, we acknowledge that this decision could lead to suboptimal flow when large gaps exist between selected sentences.

To assess this design choice, we conducted an ablation study comparing four sentence reordering strategies: 1) \textbf{Random:} randomly shuffles selected sentences, 2) \textbf{Ascending}: sorts by relevance score (low to high), 3) \textbf{Descending}: sorts by relevance score (high to low), 4) \textbf{Original order (EXIT)}: preserves sentence sequence from the source document

Table~\ref{tab:reorder} shows that maintaining the original order consistently yields the best EM and F1 scores on HQA. This confirms that preserving the source structure helps retain contextual flow, even when intermediate content is omitted. We include this result to support our design and to acknowledge the known limitations of extractive methods in handling discourse coherence.

\subsection{Training Data Ablation Analysis}

\begin{table}[]
    \centering
    \caption[Training data ablation comparison.]{Training data ablation comparison. Best results per metric are highlighted in \textbf{bold}.}
    \renewcommand{\arraystretch}{1.1}
    \small
    \label{tab:trainset_ablation}
    \begin{adjustbox}{width=0.4\textwidth}
    \begin{tabular}{l|ccc}
        \toprule[1.5pt]
        \multirow{1}{*}{\textbf{Training Data}} 
        & \scriptsize \textbf{EM} $\uparrow$
        & \scriptsize \textbf{F1} $\uparrow$
        & \scriptsize \textbf{\# token} $\downarrow$ \\
        \midrule[1pt]
        \rowcolor{gray!10} HQA & \textbf{31.6} & \textbf{42.6} & 195.1 \\
        2WIKI & 29.2 & 40.3 & \textbf{135.3} \\
        2WIKI+HQA & 30.6 & 42.0 & 232.2 \\
        \bottomrule[1.5pt]
    \end{tabular}
    \end{adjustbox}
\end{table}

Table~\ref{tab:trainset_ablation} compares models trained on HQA, 2WIKI, or both. Training solely on HQA yields the highest EM and F1 (31.6 EM, 42.6 F1) with moderate token usage. By contrast, 2WIKI training improves compression but lowers accuracy (29.2 EM, 40.3 F1). Combining datasets does not surpass HQA alone.

This finding suggests that data quality and structure matter more than quantity. HQA’s annotations appear particularly effective for learning robust compression strategies that generalize well, validating our choice to use it as the primary training dataset.

\subsection{Case Studies}

To illustrate how EXIT’s extractive compression strategy improves both accuracy and readability, we present qualitative examples and comparisons with other compression methods.

\paragraph{Original Documents vs. Ours.}  
In Table~\ref{tab:original_ours_case_study}, the original documents contain the correct answer (“Custard”) but also include distracting information (“Eggnog”). Despite having the necessary evidence, the reader fails to produce the correct answer, likely due to this distractor. By contrast, EXIT (Ours) filters out irrelevant details, drastically reduces input length, and retains only the essential context needed to answer the query accurately. As a result, the reader confidently generates the correct answer (“Custard”).

In a second scenario (Table~\ref{tab:new_case_study}), the original documents retrieve multiple documents related to “Wagner” or “sci-fi” series but fail to provide any content explicitly linking James Belushi to the correct 90s sci-fi series, resulting in an incorrect prediction. Surprisingly, EXIT removes all retrieved context entirely, providing the reader with no additional information. Under this no-context condition, the reader relies solely on its internal knowledge and, in this case, correctly identifies “Wild Palms.” While this outcome indicates a form of hallucination or model bias—since the answer emerges without external supporting evidence—it also demonstrates EXIT’s capability to avoid misleading context. By eliminating irrelevant or confusing documents, EXIT can sometimes allow the model’s internal knowledge to surface, leading to correct answers even in the absence of any retrieved information.

\paragraph{Comparisons with Other Methods.}  
Table~\ref{tab:combined_casestudy} compares EXIT with several competing compression approaches. CompAct preserves some relevant information but introduces hallucinations, causing the reader to claim that it cannot find the correct answer. Refiner omits the crucial entity required to answer the query, demonstrating how abstractive compressors may inadvertently remove key content. By contrast, EXIT avoids hallucinations and retains the answer’s entity in a concise, coherent form.

LongLLMLingua’s token-level filtering approach yields unreadable text and removes the essential “Romania” entity, preventing the reader from generating the correct answer. EXIT, on the other hand, maintains semantic coherence and includes the correct entity, allowing the reader to produce the correct answer without interference.

These case studies highlight the advantages of EXIT: it eliminates distractors, preserves critical entities, and maintains semantic integrity. Consequently, the reader consistently arrives at correct answers with fewer tokens and no hallucinations.





\newcommand{\hlc}[2][yellow]{ {\sethlcolor{#1} \hl{#2}} }

\begin{table*}[ht!]
\centering
\small
\caption{Case study comparing compressed contexts and answers between Original Docs and Ours.}
\label{tab:original_ours_case_study}
\begin{tabular}{p{0.15\textwidth} p{0.35\textwidth} p{0.35\textwidth}}
\toprule
 & \textbf{Original Docs} & \textbf{Ours} \\
\midrule
\textbf{Query} & \multicolumn{2}{p{0.7\textwidth}}{Crème Anglaise is the French version of which English dessert item?} \\
\midrule
\textbf{Context} & 
\begin{minipage}[t]{\linewidth}
[1] Crème anglaise 
\newline
Crème anglaise Crème anglaise (French for ""English cream"") is a light pouring \hlc[blue!25]{Custard} used as a dessert cream or sauce. It is a mix of sugar, egg yolks, oil, and hot milk 
\newline
...
\newline
[2] Crème anglaise
\newline
However, the ice cream base is much thicker and has various flavourings. The American South it is known as ""Custard."" It can be served like \hlc[red!25]{Eggnog} during the Christmas season. Other names include the French terms ""crème à l'anglaise"" (""English-style cream"") and ""crème française"" (""French cream""). Crème anglaise Crème anglaise (French for ""English cream"") is a light pouring Custard used as a dessert cream or sauce. It is a mix of sugar, egg yolks, oil, and hot milk often flavoured with vanilla. Its name may derive from the prevalence of sweet Custards in English desserts. The cream is made by
\newline
[3] Crème anglaise
\newline
...
\newline
[4] Custard
\newline
lemon. ""Crème pâtissière"" is a key ingredient in many French desserts including mille-feuille (or Napoleons) and filled tarts. It is also used in Italian pastry and sometimes in Boston cream pie. The thickening of the Custard is caused by the combination of egg and cornstarch. Corn flour or flour thicken at 100 °C and as such many recipes instruct the pastry cream to be boiled. In a traditional Custard such as a ""crème anglaise"", where egg is used alone as a thickener, boiling results in the over cooking and subsequent 'curdling' of the Custard; however, in a pastry cream, starch
\newline
[5] Cremeschnitte
\newline
usually pure thick Custard, less commonly combined with meringue (whipped egg whites and sugar) creme. A similar recipe with only meringue filling is called Šampita. In Australia, the dish is more commonly known as a 'vanilla slice'. Cremeschnitte A cremeschnitte (,,,,,,,, ) is a chantilly and Custard cream cake dessert popular in several Central European countries. There are many regional variations, but they all include a puff pastry base and Custard cream. In Slovenia, kremna rezina is commonly associated with the town of Bled, an Alpine tourist destination in northwestern Slovenia. The recipe
\end{minipage}
&
\begin{minipage}[t]{\linewidth}
[1] Crème anglaise 
\newline
Crème anglaise Crème anglaise (French for ""English cream"") is a light pouring \hlc[blue!25]{Custard} used as a dessert cream or sauce.
\newline
[2] Crème anglaise
\newline
Other names include the French terms ""crème à l'anglaise"" (""English-style cream"") and ""crème française"" (""French cream""). Crème anglaise Crème anglaise (French for ""English cream"") is a light pouring Custard used as a dessert cream or sauce.
\newline
[3] Crème anglaise
\newline
Alternatively, it can be drunk as a dessert on its own, for example in ""Île flottante"" (""floating island""): the cream is poured into a bowl with a piece of meringue (""blancs en neige"") floated on top along with praline. It can also be used as a base for desserts such as ice cream or crème brûlée.
\newline
[4] Custard
\newline
""Crème pâtissière"" is a key ingredient in many French desserts including mille-feuille (or Napoleons) and filled tarts.
\end{minipage} \\
\midrule
\textbf{Answer} & \multicolumn{2}{p{0.7\textwidth}}{Custard} \\
\midrule
\textbf{Predict} & 
\begin{minipage}[t]{\linewidth}
\hlc[red!25]{Eggnog}
\end{minipage}
&
\begin{minipage}[t]{\linewidth}
\hlc[blue!25]{Custard}
\end{minipage} \\
\bottomrule
\end{tabular}
\end{table*}

\begin{table*}[ht!]
\centering
\small
\caption{Case study comparing compressed contexts and answers between Original Docs and Ours. Despite containing no relevant context, Ours method predicts the correct answer, indicating a hallucination scenario.}
\label{tab:new_case_study}
\begin{tabular}{p{0.15\textwidth} p{0.35\textwidth} p{0.35\textwidth}}
\toprule
 & \textbf{Original Docs} & \textbf{Ours} \\
\midrule
\textbf{Query} & \multicolumn{2}{p{0.7\textwidth}}{Which 90s sci fi series with James Belushi was based on Bruce Wagner's comic strip of the same name?} \\
\midrule
\textbf{Context} &
\begin{minipage}[t]{\linewidth}
[1] Michael I. Wagner \newline
patterned his character after Wagner's mannerisms and physical behavior. The series ran on Thursday nights in the Spring of 1988 during the same time slot as NBC's "The Cosby Show", and with that competition could not attract a sufficient audience to get renewed for the following season.
\newline
...
\newline
Wagner helped develop and write the Bochco animated series "Capitol Critters", he also wrote and served as supervising producer
\newline
[2] Michael I. Wagner \newline
Steven Bochco and several of his projects. Wagner was asked by ABC in 1987 to help develop a new science fiction series, "\hlc[red!25]{Probe}", a light-hearted series about a scientific crime fighter named Austin James. 
\newline
...
\newline
Parker Stevenson, who played the lead character, stated in a later interview that he
\newline
[3] John Wagner \newline
the mid-1990s Wagner worked on a number of licensed properties for Dark Horse Comics in the US, including "Aliens", "Star Wars" – notably solo stories starring Boba Fett and the comics strand of the multimedia project "" – and "". 
\newline
...
\newline
It was nominated for the Angoulême International Comics Festival Prize for Scenario in 2006. In 2000 Wagner
\newline
[4] Martin Wagner (artist) \newline
Martin Wagner (artist) Martin Wagner (born April 27, 1966) is an American artist, cartoonist, and filmmaker. \newline
...
\newline. His production schedule became increasingly protracted and he ceased publishing the series altogether following issue No. 12 in 1994. In 1996 he made a
\newline
[5] Wired (film) \newline
"L.A. Law", "Murphy Brown", and "Seinfeld"), Chiklis gained fame for portraying the lead roles of Commissioner Tony Scali on the ABC police drama "The Commish" (1991-1996), and LAPD Detective Vic Mackey on the FX police drama "The Shield" (2002-2008) and as well as Marvel superhero Ben "The Thing" Grimm in the films "Fantastic Four" (2005) and "" (2007). \newline
...
\newline Wired (film) Wired is a 1989 biographical film of comedian and actor John Belushi, directed by Larry Peerce, and adapted from the 1984 book of the same name by "Washington"
\end{minipage}
&
\begin{minipage}[t]{\linewidth}
\emph{No context} (completely pruned)
\end{minipage} \\
\midrule
\textbf{Answer} & \multicolumn{2}{p{0.7\textwidth}}{Wild Palms} \\
\midrule
\textbf{Predict} &
\begin{minipage}[t]{\linewidth}
\hlc[red!25]{Probe}
\end{minipage}
&
\begin{minipage}[t]{\linewidth}
\hlc[blue!25]{Wild Palms}
\end{minipage} \\
\bottomrule
\end{tabular}
\end{table*}

\begin{table*}[ht!]
\centering
\small
\caption{Combined case studies comparing compressed contexts and answers across different methods.}
\label{tab:combined_casestudy}
\begin{tabular}{p{0.15\linewidth} p{0.35\linewidth} p{0.35\linewidth}}
\toprule
\textbf{Case Study 1} & \textbf{CompAct} & \textbf{Ours} \\
\midrule
\textbf{Query} & \multicolumn{2}{p{0.7\linewidth}}{Where was there a series of explosions and fire on 23 October 1989 (killing 23 employees and injuring 314), an explosion on 23 June 1999 (killing two men and injuring three), and an explosion and fire on 27 March 2000 (causing 1 death and 71 injuries)?} \\
\midrule
\textbf{Context} & 
The Flixborough disaster was an explosion at a chemical plant near Flixborough, England on 1 June 1974. The incident started on October 23, 1989, at 1400 Jefferson Road, \hlc[blue!25]{Pasadena, Texas}.
&
[1] Phillips disaster of 1989 \newline
On October 23, 1989, at 1400 Jefferson Road, \hlc[blue!25]{Pasadena, Texas}, a powerful explosion killed 23 and injured 314. A powerful and devastating explosion and fire ripped through the HCC, killing 23 persons—all working at the facility—and injuring 314 other (185 Phillips 66 employees and 129 contract employees).\\
\midrule
\textbf{Answer} & \multicolumn{2}{p{0.7\linewidth}}{Pasadena, Texas} \\
\midrule
\textbf{Predict} & 
I couldn't find any information about the specific incidents you mentioned. The Flixborough disaster occurred on 1 June 1974.
&
\hlc[blue!25]{Pasadena, Texas} \\
\midrule[1.5pt]

\textbf{Case Study 2} & \textbf{Refiner} & \textbf{Ours} \\
\midrule
\textbf{Query} & \multicolumn{2}{p{0.7\linewidth}}{Who said that there are only three true sports, bullfighting, mountaineering, and motor racing, the rest are merely games?} \\
\midrule
\textbf{Context} & 
1.1. Extreme sport \newline
There are only three sports: bullfighting, motor racing, and mountaineering; all the rest are merely games.
&
[1] Extreme sport \newline
usually, but wrongly, attributed to \hlc[blue!25]{Ernest Hemingway}. The phrase is: There are only three sports: bullfighting, motor racing, and mountaineering; all the rest are merely games.\\
\midrule
\textbf{Answer} & \multicolumn{2}{p{0.7\linewidth}}{Ernest Hemingway} \\ 
\midrule
\textbf{Predict} & 
There is no information provided about who said that.
&
\hlc[blue!25]{Ernest Hemingway} \\
\midrule[1.5pt]

\textbf{Case Study 3} & \textbf{LongLLMLingua} & \textbf{Ours} \\
\midrule
\textbf{Query} & \multicolumn{2}{p{0.7\linewidth}}{Which was the only Eastern Bloc country to participate in the 1984 LA Olympics?} \\
\midrule
\textbf{Context} & 
\begin{minipage}[t]{\linewidth}
\small
984 Summer Olympicscottthe majority of Soviet Bloc will not particip Games,aușescu's Roman expected.984 Summer boyc The boyc of the 1984 Summer in Los Angeles years after U.S.-led boycott of the 1980 Summer Olympics in The boycott involved 14 Bloc countries and allies, led by the Soviet Union, which initi boycott on May 8, 1984. Bootting countries organized major event, the Friendship Games, in July and August 1984 Although the boycott by the Soviet Union affected number Olympic
Sum were perhaps the first games of a new era to make a profit. Although a boycott led by the Soviet Union depleted the field in certain sports, 140 National Olympic Committees took part, which was a record at the time. Again, without the participation of the Eastern European countries, the 1984 Games were dominated by their host country. The Games were also the first time mainland \hlc[red!25]{China} (People's Republic) participated. According to British journalist Andrew Jennings, a KGB colonel stated that the agency's officers had posed as anti-doping authorities from
\end{minipage}
&
\begin{minipage}[t]{\linewidth}
\small
[1] 1984 Summer Olympics boycott \newline
The boycott involved 14 Eastern Bloc countries and allies, led by the Soviet Union, which initiated the boycott on May 8, 1984. \newline
... \newline
[3] Summer Olympic Games 
\newline
Eastern Bloc that did attend the 1984 Olympics. Although a boycott led by the Soviet Union depleted the field, 140 NOCs took part. Without Eastern European countries, the 1984 Games were dominated by the host. The Games were also the first time mainland \hlc[red!25]{China} participated. \newline
... \newline
[5] 1984 Summer Olympics boycott 
\newline
However, no threat to Eastern Bloc athletes was discovered, and the athletes from the Eastern Bloc country that did attend the 1984 games—\hlc[blue!25]{Romania}—encountered no problems.
\end{minipage} \\
\midrule
\textbf{Answer} & \multicolumn{2}{p{0.7\linewidth}}{Romania} \\
\midrule
\textbf{Predict} & 
\begin{minipage}[t]{\linewidth}
\small
\hlc[red!25]{China}
\end{minipage}
&
\begin{minipage}[t]{\linewidth}
\small
\hlc[blue!25]{Romania}
\end{minipage} \\
\bottomrule
\end{tabular}
\end{table*}

\begin{table*}[]
    \centering
    \caption[Zero-shot QA prompt evaluation of compressor performance across different Top-\(k\) scenarios, models, and datasets]{Zero-shot QA prompt evaluation of compressor performance across different Top-\(k\) scenarios, models, and datasets, measured by EM, F1, and inference latency (Lat.). 8B reader experiments were conducted on a single A100-80GB GPU, while 70B reader experiments utilized 4 A100-80GB GPUs in parallel. Best results for each dataset are highlighted in \textbf{bold}, Second best results are highlighted in \underline{underline}.}
    \label{tab:detailed_performance_comparison}
    \setlength{\tabcolsep}{4.5pt}
    \renewcommand{\arraystretch}{1.1}
    \begin{adjustbox}{width=\textwidth}
    \begin{tabular}{l|c|ccc|ccc|ccc|ccc|ccc}
        \toprule[1.5pt]
        \multirow{2}{*}{\textbf{Compressor}} & \multirow{2}{*}{\textbf{Type}}
        & \multicolumn{3}{c|}{\textbf{NQ}} 
        & \multicolumn{3}{c|}{\textbf{TQA}} 
        & \multicolumn{3}{c|}{\textbf{HQA}} 
        & \multicolumn{3}{c|}{\textbf{2WIKI}}
        & \multicolumn{3}{c}{\textbf{AVG.}} \\
        \cmidrule[0.8pt]{3-17}
        &
        & \scriptsize \textbf{EM} $\uparrow$ & \scriptsize \textbf{F1} $\uparrow$ & \scriptsize \textbf{Lat.} $\downarrow$ 
        & \scriptsize \textbf{EM} $\uparrow$ & \scriptsize \textbf{F1} $\uparrow$ & \scriptsize \textbf{Lat.} $\downarrow$ 
        & \scriptsize \textbf{EM} $\uparrow$ & \scriptsize \textbf{F1} $\uparrow$ & \scriptsize \textbf{Lat.} $\downarrow$ 
        & \scriptsize \textbf{EM} $\uparrow$ & \scriptsize \textbf{F1} $\uparrow$ & \scriptsize \textbf{Lat.} $\downarrow$
        & \scriptsize \textbf{EM} $\uparrow$ & \scriptsize \textbf{F1} $\uparrow$ & \scriptsize \textbf{Lat.} $\downarrow$ \\
        \midrule[1pt]
        \multicolumn{17}{c}{\cellcolor{gray!10}\textbf{Llama-3.1-8B-Instruct}} \\
        \midrule[0.8pt]
        \multicolumn{17}{c}{\textit{Top-5 Documents}} \\
        \midrule
        Original Docs & -
        & \underline{34.6} & \underline{47.1} & 1.0 & 58.8 & 68.6 & 0.9 & 28.1 & 38.6 & 1.0 & 16.1 & 24.9 & 1.1 & 34.4 & 44.8 & 1.0  \\
        RECOMP-Abst & Abs.
        & 31.3 & 43.2 & 1.6 & 55.9 & 65.7 & 1.4 & 26.5 & 37.0 & 2.2 & \underline{22.7} & \underline{29.1} & 2.1 & 34.1 & 43.7 & 1.8 \\
        CompAct & Abs.
        & 32.9 & 44.6 & 8.5 & 58.1 & 67.7 & 8.8 & \underline{28.8} & 39.8 & 8.3 & 16.8 & 26.0 & 8.1 & 34.2 & 44.5 & 8.4 \\
        Refiner & Abs.
        & 32.9 & 45.0 & 28.1 & 59.2 & \underline{68.9} & 10.9 & \underline{28.8} & \underline{40.0} & 6.9 & 16.8 & 25.4 & 6.4 & 34.4 & \underline{44.8} & 13.1 \\
        RECOMP-Extr & Ext.
        & \underline{34.6} & 44.6 & \textbf{0.5} & 56.5 & 65.1 & \textbf{0.4} & 23.4 & 32.8 & \textbf{0.4} & 11.2 & 19.6 & \textbf{0.6} & 31.4 & 40.5 & \textbf{0.5}  \\
        LongLLMLingua & Ext.
        & 30.2 & 41.5 & 0.9 & \underline{59.4} & 68.0 & 0.8 & 28.0 & 38 & \underline{0.8} & 21.5 & 27.4 & \underline{0.9} & \underline{34.8} & 43.7 & 0.9 \\
        
        \rowcolor{blue!5} Ours (EXIT) & Ext.
        & \textbf{35.9} & \textbf{47.8} & \underline{0.8} & \textbf{60.8} & \textbf{69.9} & \underline{0.7} & \textbf{30.6} & \textbf{41.5} & \underline{0.8} & \textbf{24.2} & \textbf{30.8} & \underline{0.9} & \textbf{37.9} & \textbf{47.5} & \underline{0.8}  \\
        \midrule
        \multicolumn{17}{c}{\textit{Top-20 Documents}} \\
        \midrule
        Original Docs & -
        & \underline{36.6} & \underline{49.5} & 3.4 & 62.0 & \underline{71.7} & 2.9 & \underline{29.9} & 40.5 & 2.9 & 18.8 & 27.9 & 3.2 & 36.8 & \underline{47.4} & 3.1  \\
        RECOMP-Abst & Abs.
        & 26.9 & 38.3 & \underline{1.7} & 57.3 & 66.6 & 1.9 & 26.8 & 37.1 & 2.4 & 22.7 & 28.8 & 2.6 & 33.4 & 42.7 & 2.2  \\
        CompAct & Abs.
        & 33.8 & 45.4 & 26.1 & 57.8 & 67.5 & 24.5 & 28.9 & 39.6 & 27.5 & 16.7 & 24.6 & 32.2 & 34.3 & 44.3 & 27.6  \\
        Refiner & Abs.
        & 30.1 & 41.4 & 28.7 & 57.6 & 67.0 & 44.2 & 26.6 & 37.3 & 29.6 & 16.3 & 24.9 & 10.8 & 32.7 & 42.6 & 28.3  \\
        RECOMP-Extr & Ext.
        & 32.8 & 42.6 & \textbf{0.6} & 55.5 & 63.6 & \textbf{0.4} & 22.2 & 31.2 & \textbf{0.5} & 10.0 & 18.3 & \textbf{0.7} & 30.1 & 38.9 & \textbf{0.6} \\
        LongLLMLingua & Ext.
        & 33.4 & 45.1 & 2.8 & \underline{62.4} & 71.2 & 2.7 & 31.2 & \underline{41.4} & 2.8 & \underline{24.1} & \underline{30.1} & 2.9 & \underline{37.8} & 46.9 & 2.8  \\
        \rowcolor{blue!5} Ours (EXIT) & Ext.
        & \textbf{38.1} & \textbf{50.8} & 1.8 & \textbf{62.8} & \textbf{72.0} & \underline{1.7} & \textbf{32.9} & \textbf{44.0} & \underline{1.8} & \textbf{25.5} & \textbf{32.3} & \underline{2.0} & \textbf{39.8} & \textbf{49.8} & \underline{1.8} \\
        \midrule[1pt]
        \multicolumn{17}{c}{\cellcolor{gray!10}\textbf{Llama-3.1-70B-Instruct}} \\
        \midrule[0.8pt]
        \multicolumn{17}{c}{\textit{Top-5 Documents}} \\
        \midrule
        Original Docs & -
        & 35.6 & \underline{48.0} & 8.6 & 65.1 & 73.9 & 7.7 & 33.7 & 44.5 & 8.3 & 20.8 & 28.3 & 9.1 & 38.8 & 48.7 & 8.4 \\
        RECOMP-Abst & Abs.
        & 34.1 & 47.0 & 4.5 & 61.3 & 70.6 & 3.3 & 30.3 & 40.8 & 4.4 & 24.2 & 30.3 & 4.2 & 37.5 & 47.2 & 4.1 \\
        CompAct & Abs.
        & 34.1 & 45.4 & 11.9 & 62.6 & 71.1 & 11.7 & 33.8 & 44.1 & 11.0 & 20.5 & 27.4 & 11.6 & 37.8 & 47.0 & 11.5 \\
        Refiner & Abs.
        & 35.3 & 47.1 & 42.5 & 64.3 & 73.0 & 18.3 & 33.8 & 44.7 & 14.6 & 21.2 & 28.0 & 11.2 & 38.7 & 48.2 & 21.6 \\
        RECOMP-Extr & Ext.
        & \underline{35.8} & 45.3 & \textbf{2.5} & 63.5 & 71.0 & \textbf{2.2} & 27.6 & 36.7 & \textbf{2.9} & 13.8 & 19.3 & \textbf{3.3} & 35.2 & 43.1 & \textbf{2.7} \\
        LongLLMLingua & Ext.
        & 32.2 & 44.0 & 4.4 & \underline{66.7} & \underline{75.2} & 3.9 & \underline{34.1} & \underline{45.3} & 4.0 & \underline{28.3} & \textbf{34.8} & 4.3 & \underline{40.3} & \underline{49.8} & 4.1 \\

        \rowcolor{blue!5} Ours (EXIT) & Ext.
        & \textbf{36.9} & \textbf{49.4} & \underline{3.9} & \textbf{67.3} & \textbf{75.9} & \underline{3.1} & \textbf{37.0} & \textbf{48.3} & \underline{3.3} & \textbf{28.6} & \underline{34.5} & \underline{3.5} & \textbf{42.5} & \textbf{52.0} & \underline{3.5} \\
        \midrule
        \multicolumn{17}{c}{\textit{Top-20 Documents}} \\
        \midrule
        Original Docs & -
        & \textbf{39.5} & \underline{52.5} & 25.8 & \textbf{69.1} & \textbf{77.6} & 24.9 & \underline{38.5} & \underline{50.0} & 25.3 & 28.8 & \textbf{36.8} & 28.1 & \underline{44.0} & \textbf{54.2} & 26 \\
        RECOMP-Abst & Abs.
        & 31.5 & 45.1 & \underline{4.5} & 63.4 & 72.2 & \underline{3.7} & 31.3 & 41.8 & 4.8 & 25.4 & 30.7 & \underline{4.8} & 37.9 & 47.4 & \underline{4.5} \\
        CompAct & Abs.
        & 33.9 & 45.1 & 30.8 & 61.7 & 70.0 & 28.1 & 31.7 & 40.9 & 32.0 & 18.5 & 23.5 & 36.5 & 36.4 & 44.9 & 31.9 \\
        Refiner & Abs.
        & 32.6 & 43.5 & 37.9 & 62.9 & 71.4 & 48.9 & 31.9 & 42.2 & 33.0 & 22.7 & 28.7 & 12.8 & 37.5 & 46.5 & 33.2 \\
        RECOMP-Extr & Ext.
        & 33.6 & 42.7 & \textbf{2.4} & 63.1 & 70.3 & \textbf{2.2} & 25.6 & 34.5 & \textbf{2.9} & 12.2 & 17.4 & \textbf{3.3} & 33.6 & 41.2 & \textbf{2.7} \\
        LongLLMLingua & Ext.
        & 34.5 & 46.4 & 10.9 & 68.2 & 76.9 & 10.4 & 36.7 & 48.4 & 10.6 & \underline{29.8} & 36.5 & 11.0 & 42.3 & 52.1 & 10.7 \\
        \rowcolor{blue!5} Ours (EXIT) & Ext.
        & \underline{39.4} & \textbf{52.6} & 5.1 & \underline{68.7} & \underline{77.3} & 4.3 & \textbf{38.6} & \textbf{50.2} & \underline{4.7} & \textbf{30.0} & \underline{36.3} & \underline{4.8} & \textbf{44.2} & \underline{54.1} & 4.7 \\
        \bottomrule[1.5pt]
    \end{tabular}
    \end{adjustbox}
\end{table*}

\begin{table*}[]
    \centering
    \caption[Few-shot QA prompt evaluation measured by EM and F1]{Few-shot QA prompt evaluation measured by EM and F1. Best results for each dataset are highlighted in \textbf{bold}, Second best results are highlighted in \underline{underline}.}
    \label{tab:performance_comparison_fewshot}
    \scriptsize  
    \setlength{\tabcolsep}{3pt} 
    \renewcommand{\arraystretch}{0.95} 
    \begin{adjustbox}{width=0.85\textwidth} 
    \begin{tabular}{l|c|cc|cc|cc|cc|cc}
        \toprule[1.5pt]
        \multirow{2}{*}{\textbf{Compressor}} & \multirow{2}{*}{\textbf{Type}}
        & \multicolumn{2}{c|}{\textbf{NQ}} 
        & \multicolumn{2}{c|}{\textbf{TQA}} 
        & \multicolumn{2}{c|}{\textbf{HQA}} 
        & \multicolumn{2}{c|}{\textbf{2WIKI}}
        & \multicolumn{2}{c}{\textbf{AVG.}} \\
        \cmidrule[0.8pt]{3-12}
        & & {\scriptsize EM$\uparrow$} & {\scriptsize F1$\uparrow$}
        & {\scriptsize EM$\uparrow$} & {\scriptsize F1$\uparrow$}
        & {\scriptsize EM$\uparrow$} & {\scriptsize F1$\uparrow$}
        & {\scriptsize EM$\uparrow$} & {\scriptsize F1$\uparrow$}
        & {\scriptsize EM$\uparrow$} & {\scriptsize F1$\uparrow$} \\
        \midrule[1pt]
        \rowcolor{gray!10}\multicolumn{12}{c}{\textbf{Llama-3.1-8B-Instruct}} \\
        \midrule[0.8pt]
        \multicolumn{12}{c}{\textit{Top-5 Documents}} \\
        \midrule
        Original Docs & - 
        & \textbf{36.9} & \textbf{48.8} & \textbf{61.5} & \textbf{70.3} & 29.6 & 39.9 & 22.2 & 29.1 & \underline{37.5} & \underline{47.0} \\
        RECOMP-Abst & Abs.
        & 33.9 & 45.1 & 57.8 & 66.7 & 27.0 & 37.2 & \textbf{26.2} & \textbf{32.2} & 36.2 & 45.3 \\
        CompAct & Abs.
        & 35.0 & 46.3 & 60.3 & 69.2 & \textbf{30.2} & \textbf{40.6} & 23.9 & 31.0 & 37.3 & 46.8 \\
        Refiner & Abs.
        & 34.4 & 46.0 & \underline{61.0} & \underline{70.0} & 29.6 & 39.9 & 23.4 & 30.0 & 37.1 & 46.5 \\
        RECOMP-Extr & Ext.
        & \underline{35.9} & 45.8 & 58.5 & 66.1 & 25.9 & 35.4 & 21.2 & 27.5 & 35.4 & 43.7 \\
        LongLLMLingua & Ext.
        & 30.7 & 41.8 & 60.8 & 68.8 & 27.3 & 37.1 & 23.0 & 28.8 & 35.5 & 44.1 \\
        \rowcolor{blue!5} EXIT (Ours) & Ext.
        & 35.8 & \underline{47.4} & \underline{61.0} & 69.8 & \underline{29.7} & \underline{40.3} & \underline{25.9} & \underline{32.0} & \textbf{38.1} & \textbf{47.4} \\
        \midrule
        \multicolumn{12}{c}{\textit{Top-20 Documents}} \\
        \midrule
        Original Docs & - 
        & \textbf{39.2} & \textbf{51.4} & \textbf{64.2} & \textbf{73.2} & 30.6 & 40.8 & 24.8 & \underline{32.4} & \underline{39.7} & \underline{49.4} \\
        RECOMP-Abst & Abs.
        & 30.2 & 40.7 & 59.3 & 67.6 & 27.4 & 37.7 & \textbf{26.8} & \underline{32.4} & 35.9 & 44.6 \\
        CompAct & Abs.
        & 35.6 & 47.0 & 60.1 & 69.2 & 30.7 & 40.6 & 21.0 & 28.3 & 36.9 & 46.3 \\
        Refiner & Abs.
        & 32.2 & 43.1 & 59.6 & 68.6 & 27.8 & 37.9 & 22.8 & 29.4 & 35.6 & 44.7 \\
        RECOMP-Extr & Ext.
        & 34.2 & 43.7 & 57.5 & 64.9 & 24.6 & 33.8 & 19.8 & 26.0 & 34.0 & 42.1 \\
        LongLLMLingua & Ext.
        & 33.8 & 45.2 & \underline{63.8} & 72.1 & \underline{31.0} & \underline{41.1} & 25.3 & 31.6 & 38.5 & 47.5 \\
        \rowcolor{blue!5} EXIT (Ours) & Ext.
        & \underline{38.8} & \underline{50.8} & 63.4 & \underline{72.3} & \textbf{32.2} & \textbf{43.1} & \underline{26.6} & \textbf{32.9} & \textbf{40.3} & \textbf{49.8} \\
        \midrule[1pt]
        \rowcolor{gray!10}\multicolumn{12}{c}{\textbf{Llama-3.1-70B-Instruct}} \\
        \midrule
        \multicolumn{12}{c}{\textit{Top-5 Documents}} \\
        \midrule
        Original Docs & -
        & \underline{39.5} & \underline{51.8} & 68.3 & 76.5 & 36.0 & 46.7 & 31.4 & 37.5 & \underline{43.8} & \underline{53.1} \\
        RECOMP-Abst & Abs.
        & 38.1 & 50.3 & 63.4 & 72.4 & 30.8 & 41.2 & 27.8 & 33.4 & 40.0 & 49.3 \\
        CompAct & Abs.
        & 37.9 & 49.7 & 67.7 & 76.0 & \underline{36.8} & \underline{47.5} & 32.2 & 38.7 & 43.7 & 53.0 \\
        Refiner & Abs.
        & 38.2 & 50.2 & 67.7 & 76.1 & 36.0 & 46.9 & 30.5 & 36.6 & 43.1 & 52.4 \\
        RECOMP-Extr & Ext.
        & \textbf{40.1} & 50.9 & 68.1 & 75.6 & 30.8 & 40.2 & 26.5 & 31.9 & 41.4 & 49.7 \\
        LongLLMLingua & Ext.
        & 35.2 & 47.2 & \underline{69.3} & \underline{77.2} & 35.6 & 46.8 & \underline{34.7} & \underline{40.2} & 43.7 & 52.8 \\
        \rowcolor{blue!5} EXIT (Ours) & Ext.
        & \underline{39.5} & \textbf{51.9} & \textbf{69.6} & \textbf{77.8} & \textbf{38.1} & \textbf{49.4} & \textbf{35.4} & \textbf{41.0} & \textbf{45.7} & \textbf{55.1} \\
        \midrule
        \multicolumn{12}{c}{\textit{Top-20 Documents}} \\
        \midrule
        Original Docs & -
        & \underline{42.1} & \underline{55.0} & \textbf{71.1} & \textbf{79.1} & \underline{39.4} & \textbf{51.1} & \underline{35.4} & \textbf{42.4} & \underline{47.0} & \textbf{56.9} \\
        RECOMP-Abst & Abs.
        & 37.2 & 50.0 & 65.6 & 74.0 & 32.4 & 43.2 & 30.3 & 35.7 & 41.4 & 50.8 \\
        CompAct & Abs.
        & 37.6 & 49.1 & 66.4 & 74.4 & 33.6 & 42.7 & 25.6 & 30.5 & 40.8 & 49.2 \\
        Refiner & Abs.
        & 36.5 & 47.6 & 66.6 & 74.7 & 33.3 & 43.3 & 30.2 & 35.6 & 41.6 & 50.3 \\
        RECOMP-Extr & Ext.
        & 38.6 & 49.1 & 68.4 & 75.6 & 28.9 & 38.0 & 24.7 & 29.9 & 40.1 & 48.2 \\
        LongLLMLingua & Ext.
        & 37.0 & 49.1 & 70.5 & 78.5 & 37.9 & \underline{49.4} & \underline{35.4} & 41.1 & 45.2 & \underline{54.6} \\
        \rowcolor{blue!5} EXIT (Ours) & Ext.
        & \textbf{42.5} & \textbf{55.3} & \underline{71.0} & \underline{79.0} & \textbf{39.8} & \textbf{51.1} & \textbf{36.5} & \underline{42.2} & \textbf{47.5} & \textbf{56.9} \\
        \bottomrule
    \end{tabular}
    \end{adjustbox}
\end{table*}

\begin{table*}[]
    \centering
    \caption[Token distribution analysis across different Top-\(k\) scenarios, models, and datasets.]{Token distribution analysis across different Top-\(k\) scenarios, models, and datasets. Best results for each dataset are highlighted in \textbf{bold}, Second best results are highlighted in \underline{underline}.}
    \label{tab:token_statistics}
    \setlength{\tabcolsep}{3pt}
    \renewcommand{\arraystretch}{1.1}
    \small
    \begin{tabular}{l|c|c|c|c|c|c}
        \toprule[1.5pt]
        \multirow{2}{*}{\textbf{Compressor}}  & \multirow{2}{*}{\textbf{Type}} 
        & \textbf{NQ} 
        & \textbf{TQA} 
        & \textbf{HQA} 
        & \textbf{2WIKI}
        & \textbf{AVG.} \\
        \cmidrule[0.8pt]{3-7}
        &
        & \scriptsize \#token (\%) \(\downarrow\)
        & \scriptsize \#token (\%) \(\downarrow\)
        & \scriptsize \#token (\%) \(\downarrow\)
        & \scriptsize \#token (\%) \(\downarrow\)
        & \scriptsize \#token (\%) \(\downarrow\) \\
        \midrule[1pt]
        \multicolumn{7}{c}{\textit{Top-5 Documents}} \\
        \midrule[0.5pt]
        Original Docs & -
        & 723.9 (100.0) & 730.3 (100.0) & 749.2 (100.0) & 784.8 (100.0) & 734.4 (100.0) \\
        RECOMP-Abst & Abs.
        & \textbf{38.0 (5.2)} & \textbf{36.9 (5.0)} & \textbf{63.3 (8.4)} & \textbf{55.1 (7.0)} & \textbf{46.0 (6.2)} \\
        CompAct & Abs.
        & 77.5 (10.7) & 79.0 (10.8) & 77.3 (10.3) & 71.4 (9.1) & 78 (10.6) \\
        Refiner & Abs.
        & 115.6 (16.0) & 103.2 (14.1) & \underline{76.6 (10.2)} & \underline{62.1 (7.9)} & 98.5 (13.4) \\
        RECOMP-Extr & Ext.
        & \underline{43.9 (6.1)} & \underline{42.7 (5.8)} & 90.2 (12.0) & 97.9 (12.5) & \underline{58.9 (8.0)} \\
        LongLLMLingua & Ext.
        & 224.3 (31) & 221.9 (30.4) & 229.2 (30.6) & 234.5 (29.9) & 225.1 (30.7) \\
        \rowcolor{blue!5} Ours (EXIT) & Ext.
        & 283.8 (39.2) & 211.3 (28.9) & 190.3 (25.4) & 154.9 (19.7) & 228.4 (31.2) \\
        \midrule[0.8pt]
        \multicolumn{7}{c}{\textit{Top-20 Documents}} \\
        \midrule[0.5pt]
        Original Docs & -
        & 2897.4 (100.0) & 2925.2 (100.0) & 2996.6 (100.0) & 3139.7 (100.0) & 2939.8 (100.0) \\
        RECOMP-Abst & Abs.
        & \textbf{26.1 (0.9)} & \textbf{38.6 (1.3)} & \textbf{64.5 (2.2)} & \textbf{51.1 (1.6)} & \textbf{43.1 (1.5)} \\
        CompAct & Abs.
        & 105.4 (3.6) & 102.5 (3.5) & 111.8 (3.7) & 109.5 (3.5) & 106.5 (3.6) \\
        Refiner & Abs.
        & 232.4 (8.0) & 176.0 (6.0) & 117.3 (3.9) & \underline{92.7 (3.0)} & 175.2 (6.0) \\
        RECOMP-Extr & Ext.
        & \underline{43.4 (1.5)} & \underline{40.6 (1.4)} & \underline{89.3 (3.0)} & 95.7 (3.0) & \underline{57.8 (2.0)} \\
        LongLLMLingua & Ext.
        & 550.9 (19.0) & 553.9 (18.9) & 563.3 (18.8) & 595.5 (19.0) & 556 (18.9) \\
        \rowcolor{blue!5} Ours (EXIT) & Ext.
        & 1001.2 (34.6) & 635.0 (21.7) & 465.0 (15.5) & 367.1 (11.7) & 700.4 (23.9) \\
        \bottomrule[1.5pt]
    \end{tabular}
\end{table*}
\begin{table*}[]
    \centering
    \caption[Latency analysis across different Top-\(k\) scenarios, models and datasets.]{Latency analysis across different Top-\(k\) scenarios, models and datasets. Each entry shows compression/reading/total time in seconds. 8B reader experiments were conducted on a single A100-80GB GPU, while 70B reader experiments utilized 4 A100-80GB GPUs in parallel. Best results for each dataset are highlighted in \textbf{bold}, Second best results are highlighted in \underline{underline}.}
    \label{tab:compression_time_comparison}
    \setlength{\tabcolsep}{4.5pt}
    \begin{adjustbox}{width=\textwidth}
    \begin{tabular}{l|c|ccc|ccc|ccc|ccc|ccc}
        \toprule[1.5pt]
        \multirow{2}{*}{\textbf{Compressor}}  & \multirow{2}{*}{\textbf{Type}}
        & \multicolumn{3}{c|}{\textbf{NQ}} 
        & \multicolumn{3}{c|}{\textbf{TQA}} 
        & \multicolumn{3}{c|}{\textbf{HQA}} 
        & \multicolumn{3}{c|}{\textbf{2WIKI}}
        & \multicolumn{3}{c}{\textbf{AVG.}} \\
        \cmidrule[0.8pt]{3-17}
        &
        & \scriptsize \textbf{Comp.} $\downarrow$ & \scriptsize \textbf{Read} $\downarrow$ & \scriptsize \textbf{Total} $\downarrow$
        & \scriptsize \textbf{Comp.} $\downarrow$ & \scriptsize \textbf{Read} $\downarrow$ & \scriptsize \textbf{Total} $\downarrow$
        & \scriptsize \textbf{Comp.} $\downarrow$ & \scriptsize \textbf{Read} $\downarrow$ & \scriptsize \textbf{Total} $\downarrow$
        & \scriptsize \textbf{Comp.} $\downarrow$ & \scriptsize \textbf{Read} $\downarrow$ & \scriptsize \textbf{Total} $\downarrow$
        & \scriptsize \textbf{Comp.} $\downarrow$ & \scriptsize \textbf{Read} $\downarrow$ & \scriptsize \textbf{Total} $\downarrow$ \\
        \midrule[1pt]
        \multicolumn{17}{c}{\cellcolor{gray!10}\textbf{Llama-3.1-8B-Instruct}} \\
        \midrule[0.8pt]
        \multicolumn{17}{c}{\textit{Top-5 Documents}} \\
        \midrule
        Original Docs & -
        & - & 1.03 & 1.03 & - & 0.93 & 0.93 & - & 0.96 & 0.96 & - & 1.12 & 1.12 & - & 1.03 & 1.03 \\
        RECOMP-Abst & Abs.
        & 1.13 & \textbf{0.43} & 1.55 & 1.06 & \textbf{0.29} & 1.35 & 1.92 & \textbf{0.32} & 2.24 & 1.73 & \textbf{0.38} & 2.11 & 1.46 & \textbf{0.36} & 1.81 \\
        CompAct & Abs.
        & 8.04 & \textbf{0.43} & 8.47 & 8.47 & 0.36 & 8.83 & 7.80 & 0.47 & 8.26 & 7.65 & \underline{0.49} & 8.14 & 7.99 & 0.44 & 8.43 \\
        Refiner & Abs.
        & 27.40 & 0.70 & 28.10 & 10.50 & 0.40 & 10.90 & 6.50 & 0.40 & 6.90 & 5.80 & 0.50 & 6.40 & 12.52 & 0.55 & 13.07 \\
        RECOMP-Extr & Ext.
        & \textbf{0.04} & \underline{0.50} & \textbf{0.54} & \textbf{0.04} & \underline{0.31} & \textbf{0.35} & \textbf{0.04} & \underline{0.37} & \textbf{0.41} & \textbf{0.04} & 0.59 & \textbf{0.63} & \textbf{0.04} & 0.44 & \textbf{0.48} \\
        LongLLMLingua & Ext.
        & 0.38 & 0.47 & 0.85 & 0.37 & 0.42 & 0.79 & 0.39 & 0.46 & 0.84 & 0.40 & 0.53 & 0.93 & 0.39 & 0.47 & 0.86 \\
        \rowcolor{blue!5} Ours (EXIT) & Ext.
        & \underline{0.33} & \textbf{0.43} & \underline{0.76} & \underline{0.35} & 0.36 & \underline{0.71} & \underline{0.38} & 0.40 & \underline{0.78} & \underline{0.39} & 0.53 & \underline{0.92} & \underline{0.36} & \underline{0.43} & \underline{0.79} \\
        \midrule
        \multicolumn{17}{c}{\textit{Top-20 Documents}} \\
        \midrule
        Original Docs & -
        & - & 3.41 & 3.41 & - & 2.85 & 2.85 & - & 2.94 & 2.94 & - & 3.22 & 3.22 & - & 3.11 & 3.11 \\
        RECOMP-Abst & Abs.
        & \underline{1.07} & 0.63 & \underline{1.70} & 1.55 & \underline{0.31} & 1.86 & 2.09 & \textbf{0.35} & 2.44 & 2.16 & \textbf{0.44} & 2.60 & 1.72 & \textbf{0.43} & 2.15 \\
        CompAct & Abs.
        & 25.58 & \underline{0.49} & 26.06 & 24.08 & 0.39 & 24.47 & 27.02 & 0.52 & 27.53 & 31.63 & 0.57 & 32.19 & 27.08 & 0.49 & 27.57 \\
        Refiner & Abs.
        & 28.00 & 0.70 & 28.70 & 43.70 & 0.50 & 44.20 & 29.00 & 0.60 & 29.60 & 9.80 & 1.00 & 10.80 & 27.63 & 0.69 & 28.32 \\
        RECOMP-Extr & Ext.
        & \textbf{0.11} & 0.51 & \textbf{0.62} & \textbf{0.11} & \textbf{0.28} & \textbf{0.39} & \textbf{0.12} & 0.42 & \textbf{0.54} & \textbf{0.11} & 0.56 & \textbf{0.68} & \textbf{0.11} & \underline{0.44} & \textbf{0.55} \\
        LongLLMLingua & Ext.
        & 1.76 & 1.04 & 2.80 & 1.78 & 0.92 & 2.70 & 1.83 & 0.93 & 2.76 & 1.89 & 1.01 & 2.90 & 1.81 & 0.98 & 2.79 \\
        \rowcolor{blue!5} Ours (EXIT) & Ext.
        & 1.33 & \textbf{0.42} & 1.76 & \underline{1.37} & 0.36 & \underline{1.74} & \underline{1.45} & \underline{0.40} & \underline{1.85} & \underline{1.51} & \underline{0.53} & \underline{2.04} & \underline{1.42} & \textbf{0.43} & \underline{1.85} \\
        \midrule[1pt]
        \multicolumn{17}{c}{\cellcolor{gray!10}\textbf{Llama-3.1-70B-Instruct}} \\
        \midrule[0.8pt]
        \multicolumn{17}{c}{\textit{Top-5 Documents}} \\
        \midrule
        Original Docs & -
        & - & 8.63 & 8.63 & - & 7.70 & 7.70 & - & 8.30 & 8.30 & - & 9.09 & 9.09 & - & 8.43 & 8.43 \\
        RECOMP-Abst & Abs.
        & 1.28 & 3.20 & 4.48 & 1.20 & \textbf{2.14} & 3.34 & 2.06 & \textbf{2.37} & 4.43 & 1.71 & \textbf{2.54} & 4.24 & 1.56 & \textbf{2.56} & 4.12 \\
        CompAct & Abs.
        & 8.77 & \underline{3.11} & 11.88 & 8.97 & 2.72 & 11.69 & 8.28 & \underline{2.73} & 11.01 & 8.36 & 3.23 & 11.59 & 8.59 & 2.95 & 11.54 \\
        Refiner & Abs.
        & 35.90 & 6.60 & 42.50 & 14.70 & 3.60 & 18.30 & 11.30 & 3.30 & 14.60 & 8.00 & 3.20 & 11.20 & 17.48 & 4.16 & 21.64 \\
        RECOMP-Extr & Ext.
        & \textbf{0.04} & \textbf{2.44} & \textbf{2.48} & \textbf{0.04} & \underline{2.21} & \textbf{2.25} & \textbf{0.04} & 2.87 & \textbf{2.91} & \textbf{0.05} & 3.29 & \textbf{3.33} & \textbf{0.04} & \underline{2.70} & \textbf{2.74} \\
        LongLLMLingua & Ext.
        & 0.50 & 3.87 & 4.37 & 0.50 & 3.40 & 3.90 & 0.51 & 3.52 & 4.03 & 0.54 & 3.72 & 4.26 & 0.51 & 3.63 & 4.14 \\
        \rowcolor{blue!5} Ours (EXIT) & Ext.
        & \underline{0.44} & 3.50 & \underline{3.94} & \underline{0.44} & 2.66 & \underline{3.10} & \underline{0.42} & 2.88 & \underline{3.30} & \underline{0.50} & \underline{3.03} & \underline{3.54} & \underline{0.45} & 3.02 & \underline{3.47} \\
        \midrule
        \multicolumn{17}{c}{\textit{Top-20 Documents}} \\
        \midrule
        Original Docs & -
        & - & 25.78 & 25.78 & - & 24.85 & 24.85 & - & 25.35 & 25.35 & - & 28.09 & 28.09 & - & 26.02 & 26.02 \\
        RECOMP-Abst & Abs.
        & \underline{1.13} & 3.36 & \underline{4.49} & \underline{1.58} & 2.11 & \underline{3.70} & 2.24 & \underline{2.59} & 4.83 & 2.07 & \textbf{2.75} & 4.81 & 1.76 & \underline{2.70} & \underline{4.46} \\
        CompAct & Abs.
        & 27.60 & \underline{3.24} & 30.84 & 25.41 & 2.74 & 28.15 & 28.90 & 3.08 & 31.99 & 33.54 & \underline{2.92} & 36.45 & 28.86 & 2.99 & 31.86 \\
        Refiner & Abs.
        & 32.50 & 5.40 & 37.90 & 47.40 & \textbf{1.50} & 48.90 & 31.40 & \textbf{1.50} & 33.00 & 9.70 & 3.10 & 12.80 & 30.25 & 2.90 & 33.15 \\
        RECOMP-Extr & Ext.
        & \textbf{0.11} & \textbf{2.27} & \textbf{2.38} & \textbf{0.12} & \underline{2.10} & \textbf{2.21} & \textbf{0.12} & 2.77 & \textbf{2.89} & \textbf{0.13} & 3.18 & \textbf{3.31} & \textbf{0.12} & \textbf{2.58} & \textbf{2.70} \\
        LongLLMLingua & Ext.
        & 2.35 & 8.51 & 10.86 & 2.35 & 8.04 & 10.38 & 2.40 & 8.24 & 10.65 & 2.50 & 8.47 & 10.97 & 2.40 & 8.32 & 10.71 \\
        \rowcolor{blue!5} Ours (EXIT) & Ext.
        & 1.62 & 3.50 & 5.12 & 1.64 & 2.67 & 4.31 & \underline{1.78} & 2.88 & \underline{4.65} & \underline{1.80} & 2.96 & \underline{4.75} & \underline{1.71} & 3.00 & 4.71 \\
        \bottomrule[1.5pt]
    \end{tabular}
    \end{adjustbox}
\end{table*}


\end{document}